\documentclass[manuscript]{jair}

\setcopyright{cc}
\copyrightyear{2025}
\acmDOI{10.1613/jair.1.xxxxx}

\JAIRAE{Insert JAIR AE Name}
\JAIRTrack{} 
\acmVolume{4}
\acmArticle{6}
\acmMonth{7}
\acmYear{2025}

\usepackage{xcolor}
\usepackage{todonotes}

\begin{document}

\title[Multi-Hypothesis Distillation]{Multi-Hypothesis Distillation of Multilingual Neural Translation Models for Low-Resource Languages}

\author{Aarón Galiano-Jiménez}
\authornote{Corresponding Author.}
\orcid{0000-0002-8107-1411}
\email{aaron.galiano@ua.es}
\affiliation{%
  \institution{Universitat d'Alacant}
  \city{Alicante}
  \country{Spain}
}

\author{Juan Antonio Pérez-Ortiz}
\orcid{0000-0001-7659-8908}
\email{japerez@ua.es}
\affiliation{%
  \institution{Universitat d'Alacant}
  \city{Alicante}
  \country{Spain}
}

\author{Felipe Sánchez-Martínez}
\orcid{0000-0002-2295-2630}
\email{fsanchez@ua.es}
\affiliation{%
  \institution{Universitat d'Alacant}
  \city{Alicante}
  \country{Spain}
}

\author{Víctor M. Sánchez-Cartagena}
\orcid{0000-0001-9600-6885}
\email{vm.sanchez@ua.es}
\affiliation{%
  \institution{Universitat d'Alacant}
  \city{Alicante}
  \country{Spain}
}


\renewcommand{\shortauthors}{Galiano et al.}

\begin{abstract}
This paper explores sequence-level knowledge distillation (KD) of multilingual pre-trained encoder-decoder translation models. We argue that the teacher model's output distribution holds valuable insights for the student, beyond the approximated mode obtained through beam search (the standard decoding method), and present Multi-Hypothesis Distillation (MHD), a sequence-level KD method that generates multiple translations for each source sentence. This provides a larger representation of the teacher model distribution and exposes the student model to a wider range of target-side prefixes.
We leverage $n$-best lists from beam search to guide the student's learning and examine alternative decoding methods to address issues like low variability and the under-representation of infrequent tokens. For low-resource languages, our research shows that while sampling methods may slightly compromise translation quality compared to beam search based approaches, they enhance the generated corpora with greater variability and lexical richness. This ultimately improves student model performance and mitigates the gender bias amplification often associated with KD.
\end{abstract}



\received{20 February 2007}
\received[revised]{12 March 2009}
\received[accepted]{5 June 2009}

\maketitle

\section{Introduction}
\label{intro}
Machine translation (MT) is an essential tool for communication and understanding among speakers of different languages. Over the past decade, the dominant architecture for MT has been the encoder-decoder Transformer~\cite{vaswani2023attention}. While encoder-decoder MT models excel in high-resource languages, they struggle with low-resource languages due to the limited availability of parallel training data~\cite{goyal-etal-2020-efficient}. This data scarcity problem becomes even more pronounced with large language models (LLMs), which has emerged as a powerful alternative to encoder-decoder models when enough training data is available. Consequently, although LLMs demonstrate strong translation performance in high-resource languages~\cite{kocmi-etal-2024-findings}, they still lag behind traditional encoder-decoder models in low-resource languages~\cite{zhu-etal-2024-multilingual, scalvini-etal-2025-rethinking, iyer-etal-2024-quality}. In this context, encoder-decoder multilingual translation models like NLLB-200~\cite{nllb}, M2M-100~\cite{fan-etal-2020-m2m}, and MADLAD-400~\cite{kudugunta2023madlad400} outperform bilingual models trained from scratch, primarily due to transfer learning from high-resource languages~\cite{facebookwmt21}. However, despite their superior performance, state-of-the-art multilingual models remain too large and computationally demanding for widespread use, especially in resource-constrained environments like laptops or smartphones.

An approach to address this challenge is knowledge distillation (KD)~\cite{hinton2015distilling}, which consists of transferring knowledge from a \textit{teacher} model to a smaller \textit{student} model. Typically, the distillation process relies on the same corpus used to train the teacher model. 
However, for most pre-trained multilingual models, this corpus is often not available. Even if the original training data were available, the knowledge of the teacher model is also derived from transfer learning across multiple languages. As a result, extracting all relevant knowledge solely from a parallel corpus may not be feasible.
In the absence of the parallel corpus used to train the teacher, one common sequence-level~\cite{kim-rush-2016-sequence} distillation strategy is to translate a monolingual text~\cite{lai-etal-2021-lmu, yu-etal-2021-hw} using beam search~\cite{graves2012sequence}, as this is the most widely used decoding method at inference time. The resulting synthetic corpus can then be used to train a student model. However, this method has several limitations. Beam search looks for the translation with the highest probability, i.e. the mode of the probability distribution~\cite{eikema-aziz-2020-map}. The mode represents a very small probability mass, and choosing a likely translation close to the mode leads to outputs with low lexical diversity~\cite{kulikov-etal-2019-importance}, over-representation of frequent tokens, and under-representation of rare tokens~\cite{muller-sennrich-2021-understanding}. In addition, it can amplify the biases present in the model, such as gender biases~\cite{vamvas-sennrich-2021-contrastive}.

We hypothesise that, contrary to the claim that knowledge comes from the top-1 prediction of the teacher~\cite{zhang-etal-2023-towards-understanding}, sampling a broader range of the model's probability distribution can yield higher quality synthetic corpora for KD. Although word-level KD~\cite{hinton2015distilling} takes advantage of this distribution, it is limited to target language prefixes present in the reference translation. To overcome these limitations, we introduce Multi-Hypothesis Distillation (MHD), a method that uses multiple translations per source sentence to provide a broader representation of the teacher's probability distribution and expose the student model to a wider range of target-side prefixes. To generate these translations, we explore different decoding methods to get the most out of the teacher model. This approach requires only monolingual corpora and supports distillation via API access, even when the teacher model itself is inaccessible. This paper evaluates MHD and analyses the key characteristics of the resulting synthetic corpora, as well as their impact on the training of student models. Our results demonstrate that MHD improves student model performance over standard sequence-level KD, even when the translations used are of lower quality than the top-1 prediction obtained with beam search, while it reduces the amplification of gender bias typically associated with KD~\cite{ahn-etal-2022-knowledge}. We also show that the choice of decoding method should consider factors such as the availability of monolingual data and the translation quality of the teacher model.\footnote{The code is available at \url{https://github.com/transducens/sampling-distillation}}\footnote{Part of this work was previously published as a Findings paper at the 2025 Annual Conference of the North America Chapter of the Association for Computational Linguistics~\cite{galiano-jimenez-etal-2025-beyond}, where we presented our initial approach. In the present work, we introduce word-level KD as a baseline, better formalise the MHD method, incorporate Minimum Bayes' Risk decoding~\cite{kumar-byrne-2004-minimum} (MBR) as an alternative decoding strategy, evaluate hallucination phenomena, and extend our overall analysis.}

The rest of the paper is organised as follows. Section \ref{releated_work} reviews related work on knowledge distillation and decoding methods. Section \ref{proposed_method} introduces our proposed MHD approach, formalising the training objective and detailing how multiple hypotheses are generated. Section \ref{settings} describes the experimental setup, including decoding methods, language pairs, corpora, and evaluation metrics. Section \ref{experiments} presents and analyses the results of our experiments, covering the impact of hypotheses number, corpus size, decoding variability, gender bias, and hallucinations. Finally, Section \ref{discussion} concludes the paper and outlines the main findings.

\section{Related work}
\label{releated_work}
There is an extensive literature on knowledge distillation and decoding methods, but the impact of decoding methods on the distillation process has been understudied. In what follows we describe the related work on KD (Sec. \ref{releated_work_kd}) and the role of decoding methods (Sec. \ref{releated_work_decoding}).

\subsection{Knowledge distillation techniques}
\label{releated_work_kd}

KD techniques can be classified into word level~\cite{hinton2015distilling} and sequence level~\cite{kim-rush-2016-sequence}. Word-level KD mimics the teacher’s probability distribution for each token, while sequence-level KD trains the student model using a synthetic corpus. This corpus is generated by the teacher by translating the source side of the original training corpus using beam search or other decoding algorithm such as Minimum Bayes' Risk decoding~\cite{kumar-byrne-2004-minimum} (MBR). In both cases, the same corpus used to train the teacher is used for the distillation. The difference is that sequence-level KD calculates the cross-entropy loss over the synthetic target side of the parallel corpus and word-level uses a combination of the cross-entropy loss over the real target and the Kullback-Leiber divergence~\cite{Kullback1951} between teacher and student probability output distributions. This means that, unlike sequence-level KD, word-level KD requires access to parallel data and greater computational resources, as both the teacher and student models must be loaded into memory simultaneously.

Regarding research on sequence-level KD with encoder-decoder multilingual translation models, some studies employ multiple teacher models, either multilingual~\cite{10.1145/3546067} or bilingual~\cite{tan2018multilingual}, to distil knowledge into a multilingual student. In contrast, our approach aims to extract as much knowledge as possible of a language pair from a single teacher in order to train a bilingual student. \citet{gumma-etal-2023-empirical} also use a single multilingual teacher, but they train a multilingual student model. Similarly, \citet{ de-gibert-etal-2023-four} distil a high-resource language pair from NLLB-200 and then fine-tune the resulting student model on a set of low-resource language pairs. Some methods use high-resource languages related to low-resource ones for distillation, training the student with both languages as sources and English as the target~\cite{song2023letz}. In contrast, our study is not limited to English as the target language.
\citet{galiano-jimenez-etal-2023-exploiting} fine-tune the teacher for specific language pairs and then train the student model with a mix of parallel and synthetic data. This method is based on parallel corpora, as well as back and forward-translation. This differs from our approach, which only uses monolingual corpora and forward-translation.

Regarding the distillation of LLMs for MT, \citet{enis2024llmnmtadvancinglowresource} used translations generated by Claude 3 Opus\footnote{\url{https://www-cdn.anthropic.com/de8ba9b01c9ab7cbabf5c33b80b7bbc618857627/Model_Card_Claude_3.pdf}} to further fine-tune an NLLB-200 1.3B model that had already been fine-tuned with translations from NLLB-200 54B and a parallel corpus. However, this additional fine-tuning did not lead to significant improvements in translation quality. In an attempt to minimise the \emph{exposure bias}~\cite{Ranzato2015SequenceLT}, \citet{agarwal2024onpolicy} combine word-level, using sequences generated by the teacher and gold references as targets, with an additional loss over the discrepancy between the teacher outputs and the student outputs. For this additional loss, they use the previous tokens generated by the student as prefixes. This approach helps the student model to learn how to generate sequences from its own predictions. Although it is a promising technique, it requires keeping the teacher and student models in memory during training, requiring more resources than our proposal.

\subsection{The role of decoding methods}
\label{releated_work_decoding}

Neural MT models generate output tokens by producing a probability distribution over the target vocabulary at each decoding step. There are two approaches for selecting these output tokens: deterministic methods, which prioritise high-probability tokens but offer low variability~\cite{kulikov-etal-2019-importance}, and stochastic methods, which sample from the probability distribution but can lead to incoherent text~\cite{basu2021mirostat}. For \textit{directed generation} tasks, such as MT, beam search~\cite{graves2012sequence} is commonly used because the output is closely tied to the input, and variability is less critical. In contrast, \textit{open-ended} tasks, like conversational chatbots, require more diverse and human-like output~\cite{holtzman2020curious}. 
Although it is common to generate the output of an LLM using sampling methods for all tasks, beam search gives better results for MT~\cite{shi-etal-2024-thorough}. While several studies analyse decoding methods~\cite{su2022contrastive, delucia-etal-2021-decoding, wiher2022decoding} and evaluate the quality of the resulting text~\cite{pillutla2021mauve}, their focus on LLMs and open-ended tasks limits their applicability to MT with encoder-decoder models.

Nevertheless, MBR~\cite{kumar-byrne-2004-minimum} has recently been used to generate sequences with an LLM to fine-tune an encoder-decoder translation model~\cite{finkelstein2024mbr} and the conclusion was that the student model fine-tuned with these sequences outperforms the model fine-tuned with beam search translations. \citet{wang2024dontthrowawaydata} extended this approach by incorporating multiple translation candidates per source sentence, similar to our proposed method. They agree with our findings in that extracting multiple sentences from the teacher model better captures its probability distribution, leading to improved student models. However, our work explores a broader range of scenarios (language pairs and different decoding methods) and concludes that the choice of decoding method should depend on both the teacher model’s translation quality and the size of the available corpus.

While the relationship between corpus quality and variability has been explored in open-ended tasks~\cite{zhang-etal-2021-trading}, it is understudied for KD in MT. However, the variability produced by different decoding methods was investigated for back-translation, and it was concluded that top-$p$~\cite{holtzman2020curious} results in higher final model performance~\cite{burchell-birch-and-kenneth-heafield-2022-exploring}.

\section{Approach: Multi-Hypothesis KD}
\label{proposed_method}
In this section we formalise the neural MT training objective based on Maximum Likelihood Estimation (MLE), the standard training objective for an MT model, and its adaptation to sequence-level KD. Then, we describe our proposed method, which generates multiple translation hypotheses per source sentence using different decoding strategies.

\subsection{Maximum Likelihood Estimation}

Given a training dataset \( \mathcal{D} = \{(x^{i}, y^{i})\}_{i=1}^{N} \), where \( x^{i} \) is a source sentence, \( y^{i} \) is its corresponding target translation and \(N\) corresponds to the number of sentences in the dataset, the training objective is to maximise the likelihood of the target sequence under the distribution of the model by minimising the following loss function:
\begin{equation}
\mathcal{L}_{\mathrm{MLE}} = -\sum_{i=1}^{N} \sum_{t=1}^{T_i} \log P(y_t^{i} \mid y_{<t}^{i}, x^{i}; \theta)
\label{loss}
\end{equation}

Where \( \theta \) represents the model parameters, \( T_i \) corresponds to the number of tokens of the target sequence, \( y_{<t} \) denotes the prefix of the $i$-th target sequence up to time step \( t-1 \) , and \(y_t^{i}\) is the target token at time step $t$.

\subsection{Sequence-Level Knowledge Distillation}


In sequence-level KD, a teacher model $\theta_{\sf T}$ generates a synthetic parallel corpus \( \mathcal{D} = \{(x^{i}, \tilde{y}^{i})\} \), where \( \tilde{y}^{i} \) is a translation generated by the model’s distribution \( P(y \mid x; \theta_{\sf T})\). The student model $\theta_{\sf S}$ is then trained to mimic the teacher's outputs by minimising the Equation \ref{loss} replacing \( y^{i}\) by  \( \tilde{y}^{i} \).

\subsection{Multi-Hypothesis Knowledge Distillation}

To increase the variety of training data we propose generating multiple translation hypotheses \( \tilde{\mathcal{Y}}_{Z}^{i} = \{\tilde{y}^{i,1}, ..., \tilde{y}^{i,M}\} \) per source sentence \( x^{i} \), where \( Z \) is the decoding method used to translate. The produced dataset is:


\begin{equation}
\mathcal{D}_{\mathrm{Z}}^{M} = \bigcup_{i=1}^{N} \bigcup_{m=1}^{M} \{(x^{i}, \tilde{y}^{i,m})\}
\end{equation}


This means that each source sentence \( x^{i} \) appears \( M \) times in the dataset, paired with different translations \( \tilde{y}^{i,m} \) generated by the teacher model using \( Z \) as the decoding method. Appendix \ref{sec:appendix_form} details the generation of $M$ translations with each decoding method. Training with this dataset results in the following loss function:

\begin{equation}
\mathcal{L}_{\mathrm{MHD}} = -\sum_{i=1}^{N} \sum_{m=1}^{M} \sum_{t=1}^{T_i} \log P(\tilde{y}_t^{i,m} \mid \tilde{y}_{<t}^{i,m}, x^{i}; \theta_{\sf S}).
\end{equation}

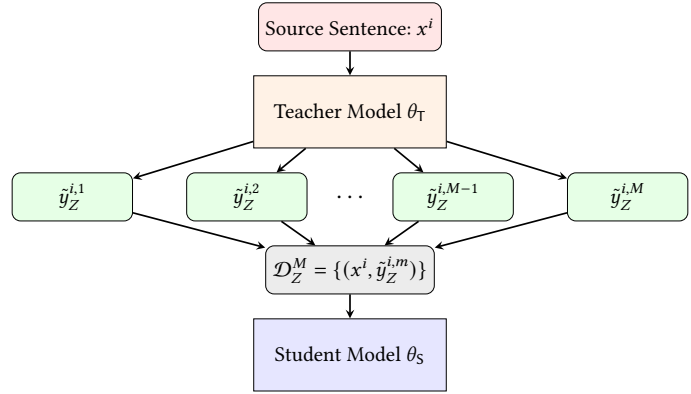
\begin{figure}[tb]
\centering
\scalebox{0.80}{
\begin{tikzpicture}[
    node distance=0.4cm and 0.1cm,
    block/.style={rectangle, draw, minimum width=2.0cm, minimum height=0.8cm, align=center, rounded corners},
    model_block/.style={rectangle, draw, minimum width=3.2cm, minimum height=1.2cm, align=center},
    evalblock/.style={block, fill=gray!15, minimum width=3.5cm},
    smallbox/.style={rectangle, draw, rounded corners, minimum width=4.5cm, minimum height=1cm},
    arrow/.style={->, thick, >=stealth},
    label/.style={font=\small}
]
\node[block, fill=red!10] (source) {Source Sentence: $x^i$};
\node[model_block, fill=orange!10, below=of source] (teacher) {Teacher Model $\theta_{\sf T}$};

\node[block, fill=green!10, below left=0.4cm and 2.0cm of teacher] (z1) {$\tilde{y}_{Z}^{i,1}$};
\node[block, fill=green!10, below left=of teacher, xshift=1.0cm] (z2) {$\tilde{y}_{Z}^{i,2}$};

\node[block, draw=none, below=of teacher] (dots) {\Large$\hdots$};

\node[block, fill=green!10, below right=of teacher, xshift=-1.0cm] (z3) {$\tilde{y}_{Z}^{i,M-1}$};
\node[block, fill=green!10, below right=0.4cm and 2.0cm of teacher] (z4) {$\tilde{y}_{Z}^{i,M}$};

\node[block, fill=gray!15, below=of dots] (dataset) {$\mathcal{D}_Z^M = \{(x^i, \tilde{y}_{Z}^{i,m})\}$};

\node[model_block, fill=blue!10, below=of dataset] (student) {Student Model $\theta_{\sf S}$};

\draw[arrow] (source) -- (teacher);
\draw[arrow] (teacher) -- (z1);
\draw[arrow] (teacher) -- (z2);
\draw[arrow] (teacher) -- (z3);
\draw[arrow] (teacher) -- (z4);

\draw[arrow] (z1) -- (dataset);
\draw[arrow] (z2) -- (dataset);
\draw[arrow] (z3) -- (dataset);
\draw[arrow] (z4) -- (dataset);

\draw[arrow] (dataset) -- (student);
\end{tikzpicture}
}
\caption{Multi-Hypothesis Knowledge Distillation (MHD) approach. The teacher model generates $M$ translations per source sentence, using $Z$ as decoding method. The resulting dataset \(\mathcal{D}_Z^M\) is then used to train the student model.}
\label{fig:mhd_pipeline}
\Description{Diagram of proposed method}
\end{figure}

Figure \ref{fig:mhd_pipeline} provides a graphical representation of this method.

\section{Experimental settings}
\label{settings}

This section describes the technical details of the experiments carried out. It starts with the decoding methods to be analysed, moves on to the language pairs, the models and the corpora used, and finally explains the features to be evaluated and how to measure each of them.

\subsection{Decoding methods}
\label{sec:decoding}

This study focuses on a selected set representing both deterministic and stochastic methods: beam search and diverse beam search as deterministic methods and top-$p$ (also known as nucleus sampling), top-$k$ and MBR decoding as stochastic methods.\footnote{We exclude ancestral sampling from our analysis because the teacher models we used in our experiments were trained using label smoothing, which elevates the likelihood of rare events, leading to translations of significantly lower quality~\cite{graca-etal-2019-generalizing}.}

\paragraph{Beam search (BS)} At each decoding step, beam search keeps the $n$ highest probability paths~\cite{graves2012sequence}. This has the advantage of identifying high probability sequences that start with less likely initial tokens and would have been ignored by greedy decoding, which always chooses the most probable token. When generating data for distillation, we used $n$=10.

\paragraph{Diverse beam search (DBS)} It is a variant of beam search that tries to produce more diverse results. Instead of maintaining a single list with the most likely paths, it divides the $n$ paths into $G$ groups and applies a penalty ($\lambda$) to prevent them from being similar to each other~\cite{diverse_beam_search}. In our experiments, as recommended by the authors of this method, we used $n$=$G$, i.e. as many groups as $n$, with only one sequence per group, and $\lambda$=$0.5$. As in the case of beam search, we used $n$=10.

\paragraph{Top-$k$} The probability mass is redistributed among the $k$ most likely tokens and the output token is then sampled for the resulting distribution~\cite{fan2018hierarchical}. This process is repeated until the end-of-sentence token is selected.\footnote{The variability of the output depends on the value of $k$. Using the vocabulary size as $k$ corresponds to ancestral sampling, while $k$=1 works as greedy decoding.}
For our experiments, we kept the original proposal of $k$=10~\cite{fan2018hierarchical}, which has proven to work well for generating synthetic corpora for back-translation~\cite{zhang2020niutrans}.

\paragraph{Top-$p$} The probability mass is redistributed among the smallest possible set of tokens whose cumulative probability exceeds the probability $p$~\cite{holtzman2020curious}. This way, the size of the set of candidate tokens can dynamically increase or decrease according to the next token's probability distribution. The sequence is generated using the same iterative process as with top-$k$. Following \citet{eikema-aziz-2022-sampling}, we set $p$ to 0.7 in our experiments.

\paragraph{Minimum Bayes' Risk decoding} MBR chooses the hypothesis that has the lowest risk of being incorrect evaluating its proximity to other candidate sequences, based on a distance measure. The method to calculate this distance is called \textit{utility function}~\cite{kumar-byrne-2004-minimum}. Following the work by \citet{finkelstein2024mbr}, we used epsilon sampling~\cite{hewitt-etal-2022-truncation} to create 256 candidates with $\epsilon$=$0.02$. We use fastChrF\footnote{fastChrF signature: numchars.6+beta.2.0+space.true}~\cite{vamvas-sennrich-2024-linear} as a utility function.

\subsection{Models, language pairs and data}

\paragraph{Models} We used NLLB-200 1.3B and NLLB-200 3.3B~\cite{nllb} as teacher models to assess the generalisation of our approach to different model sizes. 
Our students are encoder-decoder transformer models in the \emph{base} configuration, as defined by \citet[Tbl.~3]{vaswani2023attention}. With 65M parameters, our student models are notably compact, representing just 5\% of the size of the NLLB-200 1.3B model and 2\% of the NLLB-200 3.3B model. For more details on the architecture and training, see Appendix \ref{sec:appendix_A_2}.

\paragraph{Language pairs} From among NLLB-200's 200 languages, we selected language pairs based on two variables: The teacher model translation quality (Table \ref{tb:nllb_chrf_sampling}) and the size of the available corpora. Our objective is to have multiple scenarios that allow us to analyse the impact of these variables at both generation and training time. The languages we have chosen are English ({\sf eng}), Swahili ({\sf swh}), Igbo ({\sf ibo}) and Bambara ({\sf bam}).

Language pairs to be distilled and reasons for this selection are as follows:

\begin{itemize}
\setlength\itemsep{-0.1em}
    \item \textbf{From English: {\sf eng-swh}, {\sf eng-ibo}, {\sf eng-bam}}. Almost unlimited monolingual source corpora and target languages with different translation quality.
    \item \textbf{Into English: {\sf swh-eng}, {\sf ibo-eng}, {\sf bam-eng}}. Limited amount of monolingual source corpora, although in some cases sufficient to experiment with different dataset sizes. As the teacher model has been exposed to a large amount of English during its training, and BS limits the vocabulary we can extract, we hypothesise that sampling methods allow us to extract more knowledge from the teacher model.
    \item \textbf{Zero-shot: {\sf bam-swh}}. Small amount of monolingual source corpora and low translation quality. The teacher's knowledge is based on transfer learning from  other translation directions and monolingual knowledge of the source and target languages.
\end{itemize}

\paragraph{Data} 
Regarding the availability of data, English and Swahili have the largest corpora, from which we selected a subset of $1$~million monolingual sentences. For Igbo, we used a corpus comprising 451,789 sentences, while for Bambara we employed a corpus containing 108,187 sentences. All corpora used are freely available. Table \ref{tab:corpora_stats} shows the origin of each corpus. Specific details about the corpora and their pre-processing can be found in Appendix \ref{sec:appendix_A_1}.

As development and test sets we use the FLORES+\footnote{\url{https://github.com/openlanguagedata/flores}}~\cite{nllb} \textit{dev} and \textit{devtest} splits, respectively.

\begin{table}[tb]
\centering
\resizebox{\columnwidth}{!}{
\begin{tabular}{lccccccc}
\hline
\small{\textbf{Method}} & \small{\textbf{eng-swh}} & \small{\textbf{eng-ibo}} & \small{\textbf{eng-bam}} & \small{\textbf{swh-eng}} & \small{\textbf{ibo-eng}} & \small{\textbf{bam-eng}} & \small{\textbf{bam-swh}} \\
\hline
\small{BS} & \small{$59.2$} & \small{$41.0$} & \small{$30.9$} & \small{$63.5$} & \small{$52.6$} & \small{$38.6$} & \small{$35.6$} \\
\small{DBS} & \small{$58.5$} & \small{$39.0$} & \small{$28.6$} & \small{$63.0$} & \small{$51.7$} & \small{$37.6$} & \small{$32.6$} \\
\small{top-$p$} & \small{$51.3$} & \small{$36.3$} & \small{$27.0$} & \small{$57.0$} & \small{$46.7$} & \small{$35.5$} & \small{$32.6$} \\
\small{top-$k$} & \small{$49.1$} & \small{$34.4$} & \small{$27.2$} & \small{$52.3$} & \small{$44.3$} & \small{$34.7$} & \small{$32.5$} \\
\small{MBR} & \small{$58.9$} & \small{$41.8$} & \small{$31.7$} & \small{$63.8$} & \small{$53.0$} & \small{$38.7$} & \small{$36.6$} \\
\hline
\end{tabular}
}
\caption{ChrfF++ scores of NLLB-200 1.3B on the FLORES+ devtest dataset for different decoding methods: beam search (BS), diverse beam search (DBS), top-$p$ (average of 3 runs), top-$k$ (average of 3 runs), and MBR. The results with BLEU are showed in Appendix \ref{sec:nllb_1_BLEU}. The NLLB-200 3.3B results, which show the same relative order between decoding methods, can be found in Appendix \ref{nllb_3_methods}.}
\label{tb:nllb_chrf_sampling}
\end{table}

\begin{table}[tb]
    \centering
    \resizebox{\columnwidth}{!}{
    \begin{tabular}{clr}
        \hline
        \textbf{Language} & \textbf{Corpus} & \textbf{Sentences} \\
        \hline
        English & OSCAR-3301 
        & 1,000,000 \\
        Swahili & Monolingual African Languages from ParaCrawls 
        & 1,000,000 \\
        Igbo & Monolingual African Languages from ParaCrawls 
        & 451,789 \\
        Bambara & bayelemabaga, lafand-mt, Leipzig, NLLB-Seed, xP3, MADLAD-400 & 108,187 \\
        \hline
    \end{tabular}
    }
    \caption{Monolingual corpora used.}
    \label{tab:corpora_stats}
\end{table}

\subsection{Evaluation metrics}
We evaluate two key elements: the synthetic corpora and the models trained on them.
\paragraph{Synthetic corpora} We assess vocabulary diversity, hypotheses variability, and translation performance of the teacher model used for its generation.

\begin{itemize}
    \item Lexical richness: measured through Zipf's Law and by counting unique words and sentences.
    \item Variability among the $M$ translations generated for each source: evaluated using self-BLEU~\cite{10.1145/3209978.3210080}, where lower values indicate greater diversity in translations from the same source sentence.
    \item Translation performance of the teacher model: we evaluate translation performance on the FLORES+ dataset, generating a single translation per source. We use chrF++~\cite{popovic-2017-chrf} to estimate the translation quality of the synthetic corpora generated by the teacher model for each language pair. The original NLLB paper reports that chrF++ scores for this model tend to be systematically higher when generating English compared to other target languages, and that these scores correlate well with human evaluations~\cite{nllb}. This observation supports the use of chrF++ as a reliable proxy for synthetic corpus quality. 
\end{itemize}

All synthetic corpora are generated by translating the corresponding monolingual corpus (see Table \ref{tab:corpora_stats}) with the teacher model, using the Transformers library~\cite{wolf-etal-2020-transformers} and the selected decoding method.

\paragraph{Student models} We assess student performance based on four criteria:

\begin{itemize}
    \item Translation performance: evaluated in the same manner as the teacher's output, using beam search ($n$=5) on the test set. Although recent neural evaluation metrics such as AfriCOMET~\cite{wang-etal-2024-afrimte} and SSA-COMET~\cite{li2025ssacometllmsoutperformlearned} support some of the languages considered in this work, none of these metrics have been trained specifically to cover the full set of language pairs under study. Given this limitation, we adopt chrF++\footnote{chrF++: "nrefs:1|case:mixed|eff:yes|nc:6|nw:2|space:no|}~\cite{popovic-2017-chrf} as our primary evaluation metric. To validate our findings further and provide a more comprehensive assessment, we also report BLEU\footnote{BLEU: "nrefs:1|case:mixed|eff:no|tok:13a|smooth:exp|} and, for language pairs that are supported ({\sf eng-swh} and {\sf swh-eng}), COMET~\cite{rei-etal-2020-comet} scores. We observe a consistent ranking of systems in most scenarios, indicating that the reported trends are stable and not tied to a specific evaluation method.
    \item Statistical significance: using paired approximate randomization~\cite{riezler-maxwell-2005-pitfalls} to compare the student models trained with different parallel corpora \(\mathcal{D}_{\mathrm{Z}}^{M}\).
    \item Gender bias: measured through contrastive conditioning~\cite{vamvas-sennrich-2021-contrastive}, as detailed in Sec.~\ref{bias}, due to the lack of annotated datasets for the languages of this study.
    \item Hallucinations: assessed by calculating the cosine similarity between the sentence embeddings (see Sec.~\ref{hallucinations}) of the generated translations and the reference.
\end{itemize}

\section{Experiments and results}
\label{experiments}

Our experiments are designed to evaluate the effectiveness of MHD and to detect the factors influencing student model performance. First, in Section \ref{sec:100k_sampling}, we evaluate MHD using small source corpora and varying the number of translation hypotheses per sentence ($M$). The aim is to understand how exposing the student model to multiple translations and target-side prefixes affects learning. In this setting, our baseline is the student model distilled with word-level KD and \(\mathcal{D}_{\mathrm{BS}}^{1}\) as a parallel corpus, allowing us to compare the effect of directly transferring the teacher's token-level distribution against our sequence-level approach.

Once the best-performing value of $M$ is identified, we fix it and study the impact of increasing the size of the monolingual source corpus (Section \ref{sec:corpus_size}). This allows us to assess how the amount of data influences the diversity of vocabulary and sentence structures available for knowledge extraction from the teacher. These experiments are conducted only for language pairs for which we have access to larger monolingual corpora.

Next, in Section \ref{sec:trade-off}, we explore the trade-off between diversity and translation quality introduced by the decoding parameters $p$ (in top-$p$) and $k$ (in top-$k$), which control how far the sampled outputs diverge from the teacher model's most likely predictions.
We then conducted a gender bias analysis (Section \ref{bias}) and, finally, in Section \ref{hallucinations}, we examine the tendency to hallucinate of our student models.


\subsection{Impact of the number of translation hypotheses and decoding methods}
\label{sec:100k_sampling}

Sampling methods typically yield lower performance for MT compared to BS and DBS as shown in Table~\ref{tb:nllb_chrf_sampling}. Nevertheless, they are widely used in open-ended generation tasks because of their ability to produce diverse and more human-like outputs than deterministic methods. In this section, we investigate whether, despite the drop in translation quality, sampling methods can provide more effective training data for student models.

We generated our \(\mathcal{D}_{Z}^{M}\) datasets by translating 100k sentences (as this is the size of our smallest corpus) using \(Z\)=\{BS, DBS, \text{top-$p$}, \text{top-$k$}, MBR\} and \(M=\{1, 3, 5, 10\}\). As already explained in Section \ref{sec:decoding}, while top-$p$ and top-$k$ rely on sampling, BS and DBS selected the \(M\) highest-probability candidates. For MBR, we selected the best \(M\) translations. Note that \(\mathcal{D}_{\mathrm{BS}}^{1}\) corresponds to the standard sequence-level KD. Subsequently, we trained student models on the training datasets generated with each decoding method.

\begin{figure}[tb]
    \centering
    \includegraphics[scale=0.34]{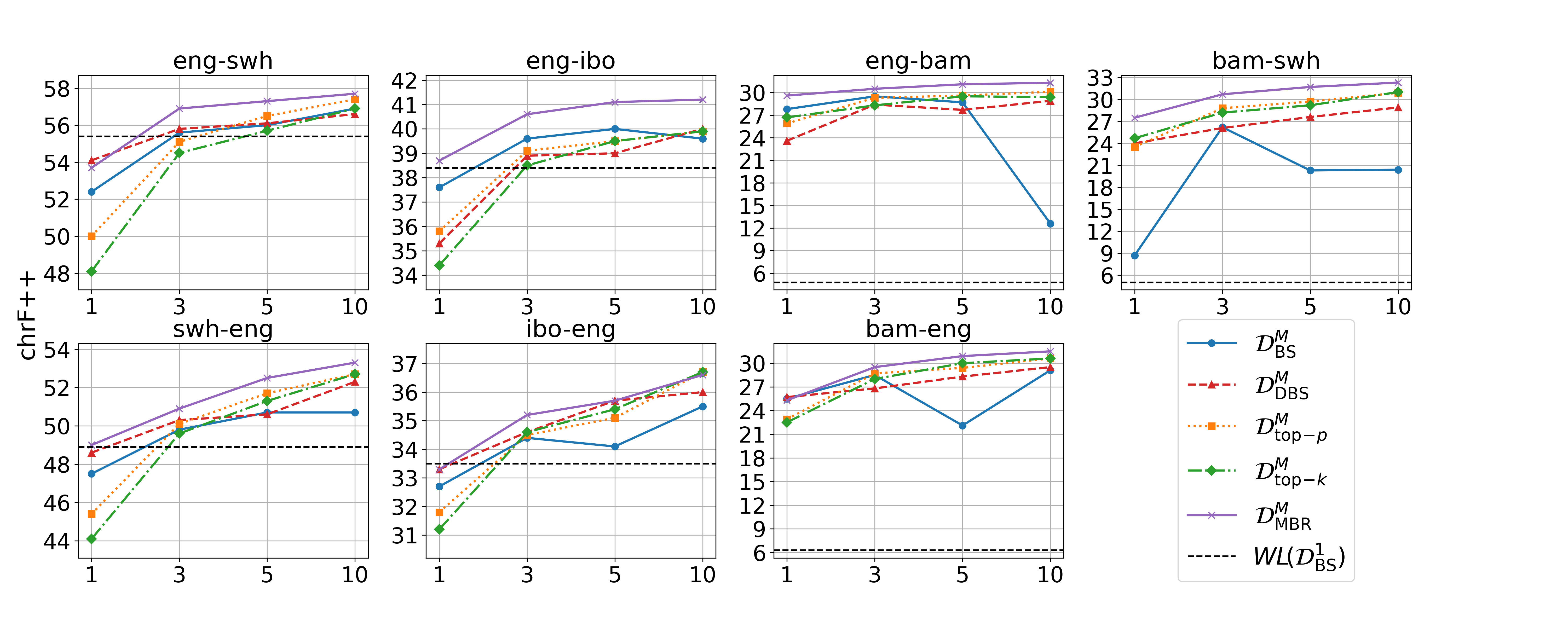}
    \caption{Average chrF++ score obtained by student models trained on M samples generated with different decoding methods (x-axis).}  
    \label{fg:chrf_student}
    \Description{MHD results with 100k monolingual sentences and different M values.}
\end{figure}

\paragraph{Results} Fig.~\ref{fg:chrf_student} shows the performance of student models distilled from NLLB-200 1.3B, with scores reflecting the average chrF++ on three training runs.\footnote{Note that, for the sampling methods, translations were  sampled again from the teacher's distribution in each training run.} Results with NLLB-200 3.3B, which shows the same relative order between decoding methods, are provided in Appendix \ref{sec:appendix_B_1}.

The results of the statistical significance tests are shown in Appendix \ref{sec:appendix_B_2}, Table \ref{tb:bleu_1M}. To examine the differences between MHD and standard sequence-level KD, we first compared the student models trained on \(\mathcal{D}_{\mathrm{Z}}^{10}\) (with \(Z\)=\{BS, DBS, \text{top-$p$}, \text{top-$k$}, MBR\}) with the student model trained on \(\mathcal{D}_{\mathrm{BS}}^{1}\). Then, to assess the impact of different decoding methods, we compared the models trained on each \(\mathcal{D}_{\mathrm{Z}}^{10}\) to those trained on \(\mathcal{D}_{\mathrm{BS}}^{10}\). Except for the {\sf eng-ibo} models, all language pairs showed statistically significant differences compared to \(\mathcal{D}_{\mathrm{BS}}^{1}\). In contrast, when considering \(\mathcal{D}_{\mathrm{BS}}^{10}\) as the baseline, models trained with \(\mathcal{D}_{\mathrm{DBS}}^{10}\) for {\sf ibo-eng}, {\sf bam-eng}, and {\sf eng-swh} did not show differences. Models trained on datasets generated by sampling methods showed statistically significant differences for {\sf eng-bam}, {\sf swh-eng}, {\sf ibo-eng}, {\sf bam-eng} and {\sf bam-swh}.

Word-level KD (\(\mathrm{WL}(\mathcal{D}_{\mathrm{BS}}^{1})\) in Figure \ref{fg:chrf_student}) yields superior results compared to sequence-level KD with a single sample for several language pairs. This suggests that, when the teacher model is well-calibrated, its probability distribution contains valuable information for training student models. However, if the teacher model is poorly calibrated for a specific language, its probability distribution can exhibit high entropy, proving detrimental to the student model. This phenomenon is observable in language pairs involving Bambara. MHD allows for the extraction of more information regarding the teacher's probability distribution than traditional sequence-level methods. Crucially, due to the specific decoding methods employed, this distribution is constrained to a subset of tokens rather than the entire vocabulary, thereby mitigating issues associated with highly dispersed probability mass. This benefit, coupled with the generation of multiple prefixes, enables MHD to outperform word-level KD and standard sequence-level KD across all investigated language pairs.

As expected, student models trained on deterministic methods or MBR outputs generally performed better than top-$p$ and top-$k$ when only one translation per sentence was generated ($M=1$). However, as the number of translations per sentence increased, models trained on sampled data outperformed those trained with deterministic methods.

The gap between BS and sampling methods is especially notable for {\sf bam-swh}. Interestingly, students trained with \(\mathcal{D}_{\mathrm{BS}}^{5}\) and \(\mathcal{D}_{\mathrm{BS}}^{10}\) performed worse than those trained with \(\mathcal{D}_{\mathrm{BS}}^{3}\). \citet{eikema-aziz-2020-map}'s observations on the inadequacy of the mode show that, when the model is poorly fitted, the most probable translation is not the best. This can explain why traditional KD with BS (\(\mathcal{D}_{\mathrm{BS}}^{1}\)) fails, but \(\mathcal{D}_{\mathrm{BS}}^{3}\) contains translations from which the student model is able to learn. The results with \(\mathcal{D}_{\mathrm{BS}}^{M>3}\) are discussed further on, together with the analysis of the generated corpora.

\(\mathcal{D}_{\mathrm{MBR}}^{M}\) achieved the best results for all tested values of $M$ in language pairs involving Bambara, but did not outperform the other methods for the other languages. We hypothesise that it is for these language pairs ({\sf bam-eng}, {\sf eng-bam}, {\sf bam-swh}) that the mode is most inappropriate. This is where extracting more information from the probability distribution is most beneficial; MBR helps to extract that information while filtering out possible mistranslations. Nevertheless, we cannot rule out a metric bias introduced by using the same type of metric to rank the MBR candidates and for evaluation~\cite{kovacs-etal-2024-mitigating}, given that MBR is not the most effective method when evaluated using BLEU (Appendix \ref{sec:appendix_B_1}).

In general, with \(M = 10\), the greatest difference between sampling and deterministic methods occurs when the target language is English. This is in line with our hypothesis that, as the teacher has been trained with a vast amount of {\sf eng} and we are working with small corpora, the sampling methods allow us to extract more information than BS.

To ensure that the improvements seen with sampling methods were not simply due to a particularly good translation among the multiple outputs, we conducted an additional experiment. For the {\sf eng-swh} \(\mathcal{D}_{\text{top-$p$}}^{10}\) corpus, we selected the best translation for each source sentence based on COMET without reference. We then used only these selected translations to train a student model. The resulting performance was similar to that of a model trained with \(\mathcal{D}_{\text{top-$p$}}^{1}\), confirming that the improvements observed with sampling methods were driven by the diversity of multiple translations.

\paragraph{Analysis of generated corpora} To explain the above results, we analyse the properties of each decoding method, as well as the corpora that were generated.

\begin{figure}[tb]
    \centering
    \includegraphics[scale=0.35]{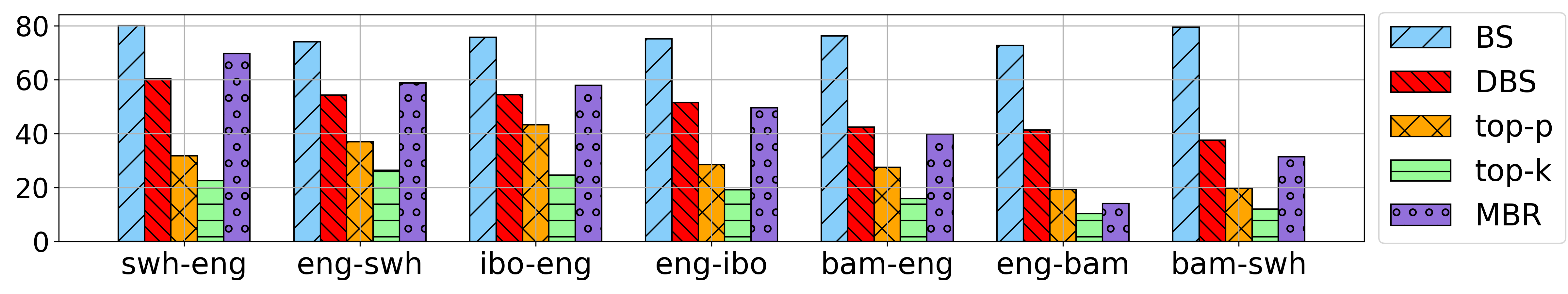}
    \caption{Similarity among 10 generated translations per source sentence as evaluated by self-BLEU (y-axis).}
    \label{fg:self-bleu}
    \Description{Self-BLEU scores of 10 translations from the same source sentence for each decoding method.}
\end{figure}

\begin{figure}[tb]
    \centering
    \hspace*{-0.6cm}\includegraphics[scale=0.30]{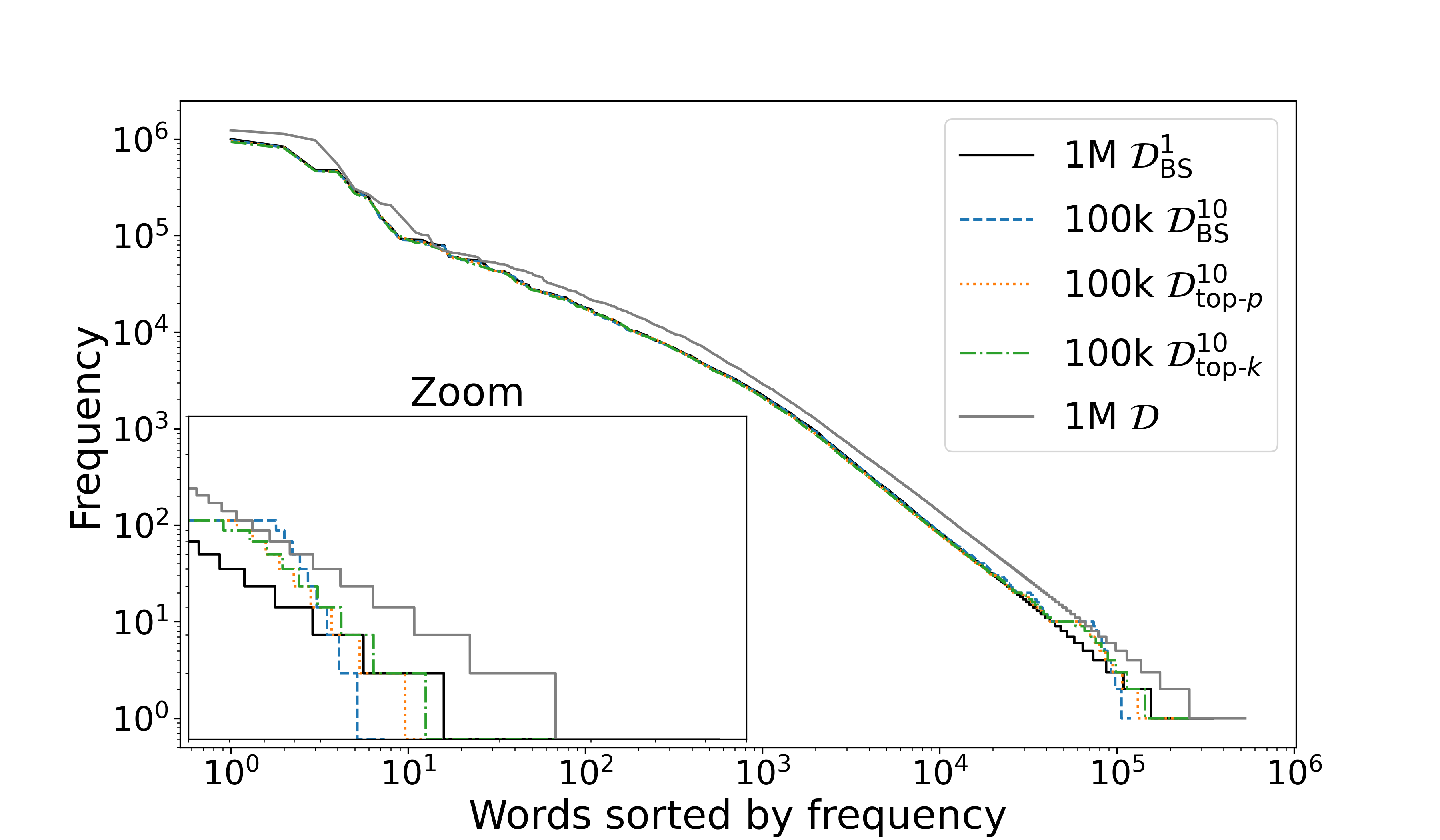}
    \caption{Zipf's distribution over Swahili corpora. Similar patterns were observed for the other languages.}
    \Description{Swahili Zipf's law.}
    \label{fg:zipf_2}
\end{figure}

The variability of the generated translations indicates the amount of information extracted. Figure \ref{fg:self-bleu} shows the self-BLEU~\cite{10.1145/3209978.3210080} score of 10 translations per source generated using each decoding method. As self-BLEU measures the similarity between translations, a high score indicates low variability. As expected, deterministic methods, which focus on selecting the most likely translations, result in low variability, even when using DBS. In contrast, sampling methods, especially top-$k$, produce more diverse translations. This variability suggests that the translations generated using sampling methods have a greater vocabulary and are richer in terms of lexicon. This explains why they provide better training data for the student models. To support this claim, we use the Zipf distribution~\cite{holtzman2020curious} of each generated corpus. Figure \ref{fg:zipf_2} compares the Zipf distribution of the corpora generated by translating 100k sentences from English into Swahili, together with the distribution of a native Swahili corpus of 1M sentences (1M \(\mathcal{D}\) in the plot). The Figure also includes the distribution of a corpus generated by translating the English corpus of 1M sentences with BS and $M$=1. The analysis shows that the corpora generated with sampling and $M$=10 from smaller texts are more similar to native corpora than those produced with BS with a single translation from larger texts. Intuitively, the generation of multiple translations with BS might result in either very similar sentences, adding little value, or hallucinations when the model is forced into less probable paths.

\begin{figure*}[tb]
    \centering
    \includegraphics[scale=0.30]{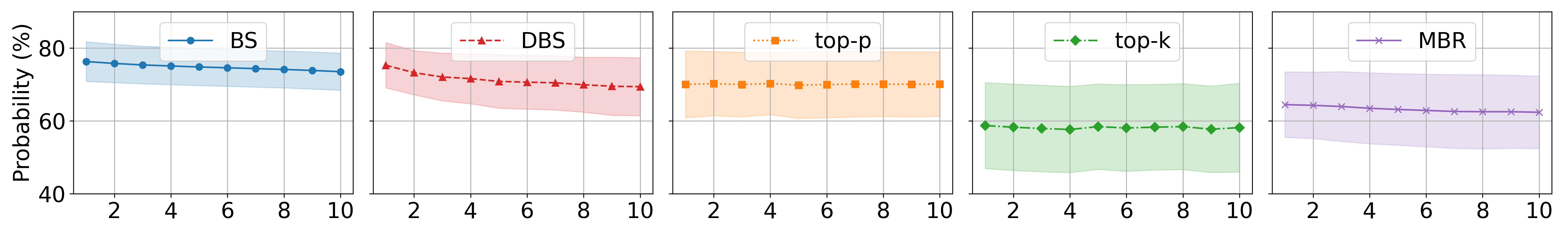}
    \caption{Probabilities normalised by length of 10 {\sf swh-eng} translation hypotheses generated with NLLB-200 1.3B for each source sentence in the FLORES+ {\sf devtest} set. The shaded areas around each line represent the standard deviation.}
    \label{fg:probabilities}
    \Description{Probabilities generated by each decoding method.}
\end{figure*}

The idea that unlikely paths lead to poorer translations is related to how each decoding method generates its translations. While sampling methods generate translations independently, deterministic methods and MBR perform a ranking of the generated hypotheses. As seen in Figure \ref{fg:probabilities}, this results in the probabilities of top-$p$ and top-$k$ translations remaining stable, while those of deterministic methods and MBR decay. This may explain the decrease in quality observed in students trained with \(\mathcal{D}_{\mathrm{BS}}^{M>1}\) when working with languages for which the teacher is poorly fitted. If the next-token probability mass is highly concentrated in a few tokens and we require more translations, deterministic methods have to take unprovable paths.

\begin{figure}[tb]
    \centering
    \includegraphics[scale=0.30]{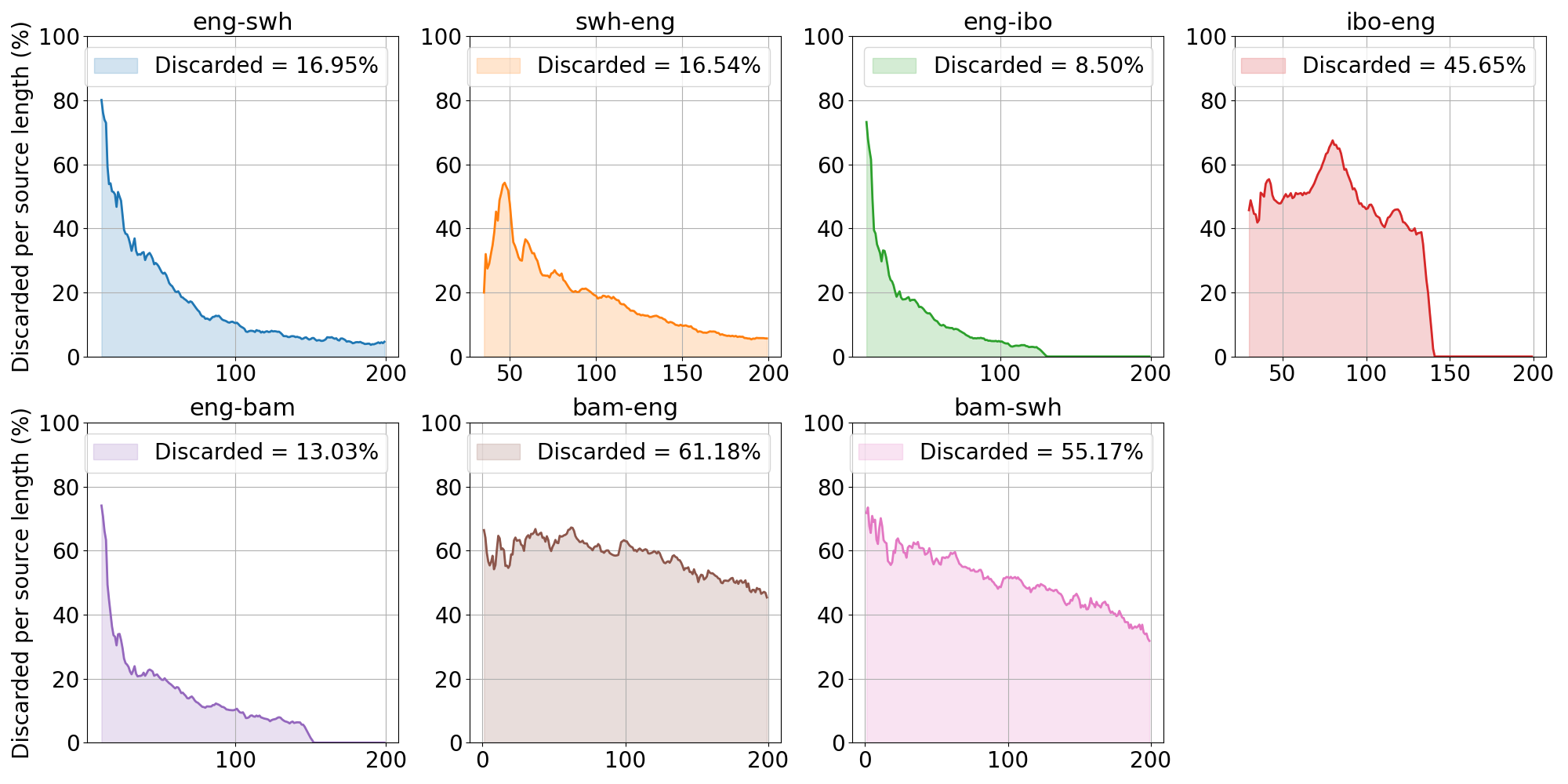}
    \caption{Percentage of translations (y-axis) for the same source with a probability lower than the median of the 256 translations generated using epsilon sampling for the same source. Source sentences are ordered by their length in characters (limited to 200) along the x-axis. The shaded area represents the total percentage of translations that were discarded.}
    \label{fg:filter}
    \Description{Filtered sentences using the expectation.}
\end{figure}

To further investigate this phenomenon, we conducted an additional experiment to approximate the expected quality of the teacher distribution. Using the 256 translations per source sentence generated with epsilon sampling, we computed the median probability of these translations as a proxy for the expected translation likelihood. Then, we filtered them by removing those with a probability lower than the previously obtained median. This allowed us to identify which beam outputs were statistically unlikely under a broader sampling of the teacher distribution. The results, illustrated in Figure~\ref{fg:filter}, show that the discarded translations were predominantly associated with short source sentences and language pairs where the teacher model performed poorly. This supports the hypothesis that deterministic methods, when forced to produce multiple outputs, are more likely to select low-probability and potentially low-quality continuations in such settings. 


In contrast to deterministic methods, sampling methods can produce repeated sentences, especially top-$p$ due to its dynamically adjusted window size. While this can limit diversity, it can also help to prevent hallucinations that could negatively affect the training of student models. This effect is visible in Figure~\ref{fg:self-bleu}, where top-$p$ sampling produces lower variability than top-$k$.


The performance of MBR depends on the probability distribution of the teacher model. When the probability mass is highly concentrated in a few tokens, MBR produces low variability (Figure \ref{fg:self-bleu}, {\sf swh-eng}). Conversely, when the teacher's probability distribution is more spread out, MBR introduces greater diversity in its selections (Figure \ref{fg:self-bleu}, {\sf eng-bam}).


\subsection{Impact of source corpus size}
\label{sec:corpus_size}

The size of the source corpus plays a crucial role in KD, as a larger corpus contains more vocabulary and allows for more knowledge to be extracted from the teacher. To analyse how this affects MHD, we translated 100k, 500k and 1~million sentences, with $M$=10. 
We used \(\mathcal{D}_{\mathrm{BS}}^{1}\) obtained from the same corpus as a baseline for each size.

\begin{figure}[tb]
    \centering
    \includegraphics[scale=0.34]{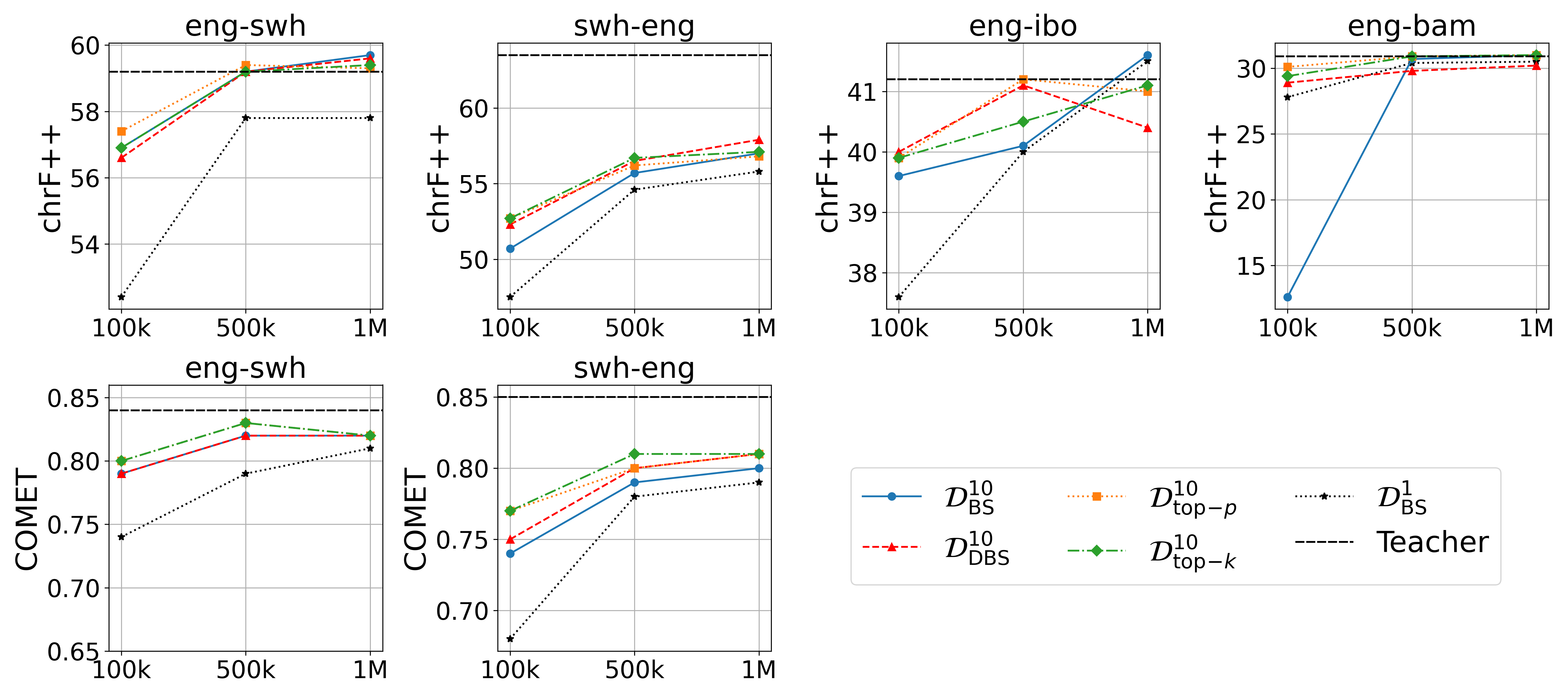}
    \caption{Scores attained for different corpus sizes in number of sentences (x-axis). \(\mathcal{D}_{\mathrm{BS}}^{1}\)  corresponds to the standard sequence-level KD. The results of \(\mathcal{D}_{\text{top-$p$}}^{10}\) and \(\mathcal{D}_{\text{top-$k$}}^{10}\) overlap in the COMET {\sf eng-swh} graph, as well as \(\mathcal{D}_{\mathrm{BS}}^{10}\) and \(\mathcal{D}_{\mathrm{DBS}}^{10}\)}
    \label{fg:bleu_student_100k-1M}
    \Description{...}
\end{figure}

\paragraph{Results} Figure \ref{fg:bleu_student_100k-1M} shows the performance of student models trained on each dataset. As observed, the discrepancy between different decoding methods decreases as the corpus size increases. For corpora of 500k sentences, sampling methods still outperform BS, while for corpora of 1~million sentences, sampling methods do not consistently yield superior results. However, MHD remains advantageous over \(\mathcal{D}_{\mathrm{BS}}^{1}\).

\begin{figure}[tb]
    \centering
    \includegraphics[scale=0.34]{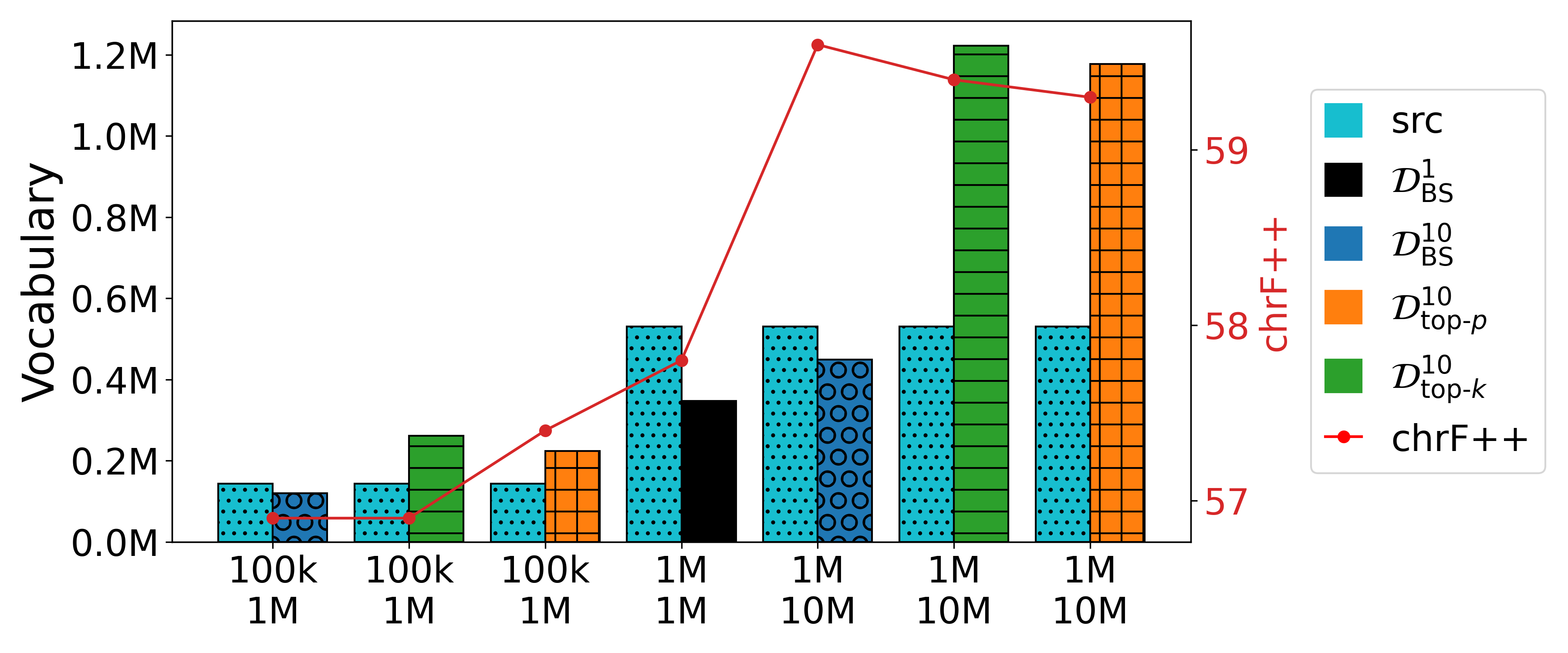}
    \caption{Relationship between the vocabulary size of the {\sf swh-eng}) training corpus and the chrF++ of the student models. X-axis markers indicate amount of sentences in the source corpus (first row) and sentences in the generated corpus (second row).}
    \label{fg:vocabulary-bleu}
    \Description{...}
\end{figure}

\begin{figure}[tb]
    \centering
    \includegraphics[scale=0.32]{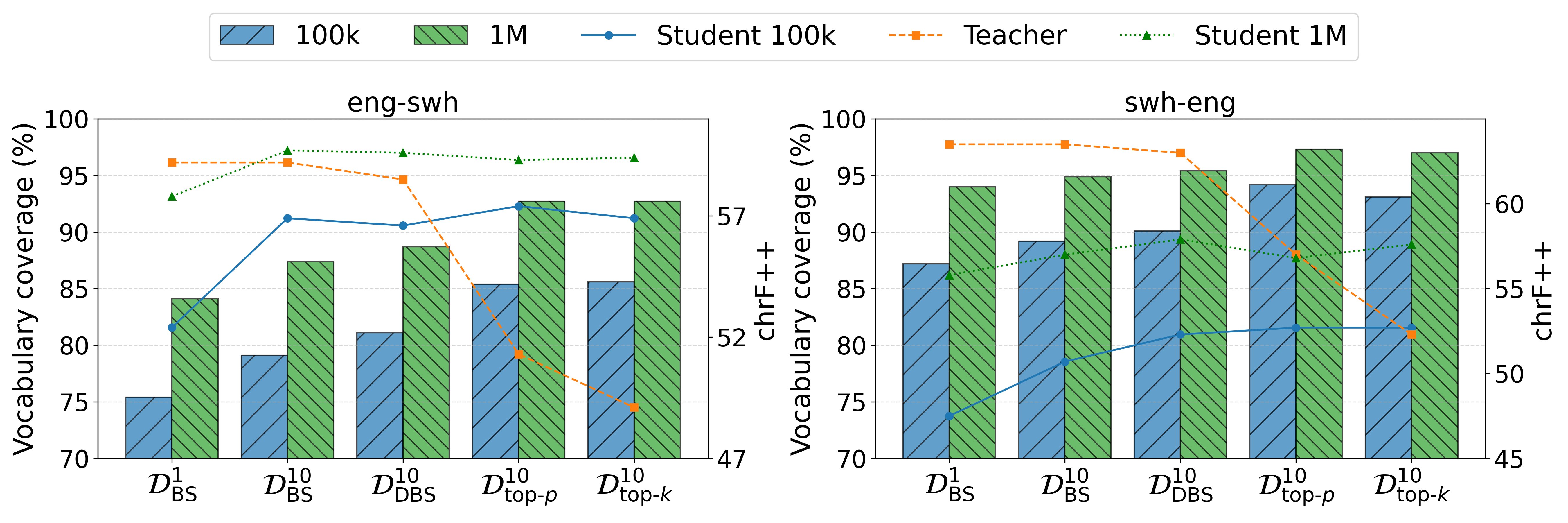}
    \caption{Effect of vocabulary coverage and teacher translation quality. X-axis shows decoding methods ranked by variability. Columns show the percentage of {\sf devtest} vocabulary (left Y-axis) present in the training corpus. Lines show the chrF++ (right Y-axis) of the models trained with each corpus and the chrF++ of the teacher generating only one translation per source sentence with the decoding method used to generate each dataset.}
    \label{fg:vocabulary-coverage.}
\end{figure}

\paragraph{Analysis of generated corpora} To explore the importance of lexical richness in the translated corpus, we compared the number of unique words in both the source corpus and the generated corpus. Figure \ref{fg:vocabulary-bleu} illustrates the relationship between the size of the source vocabulary, the vocabulary produced by each decoding method, and the chrF++ scores obtained by training student models on these corpora. The results show that sampling methods act as vocabulary amplifiers by generating multiple translations. 
However, it is important to know which part of this vocabulary is useful to the model. Figure \ref{fg:vocabulary-coverage.} shows the percentage of the {\sf devtest} target vocabulary present in the training corpus. It can be seen that until a certain coverage is reached (about 87\% for {\sf eng-swh} and 95\% for {\sf swh-eng}), increasing the coverage produces better student models, even if the teacher translations are worse. On the other hand, once this threshold is exceeded, it is more beneficial to prioritise translation quality.

In addition to translation quality, BS can offer another benefit for KD. During training, models typically use teacher forcing, where the correct token is used as input, leading to a mismatch between training and inference. During inference, the model must rely on self-generated tokens, typically obtained with BS. This \emph{exposure bias}~\cite{Ranzato2015SequenceLT} can be mitigated by training the student models with BS outputs, which are closer to the tokens generated during inference. If the source corpus is sufficiently large, BS can extract enough vocabulary, and its similarity to the inference process benefits the student model. This also explains the performance of the {\sf swh-eng} model trained on a 1M source corpus using DBS (Figure \ref{fg:bleu_student_100k-1M}), which keeps the inference similarity of BS while providing greater diversity.

\subsection{Divergence from the mode vs. translation quality}
\label{sec:trade-off}

\begin{figure}[tb]
    \centering
    \includegraphics[scale=0.38]{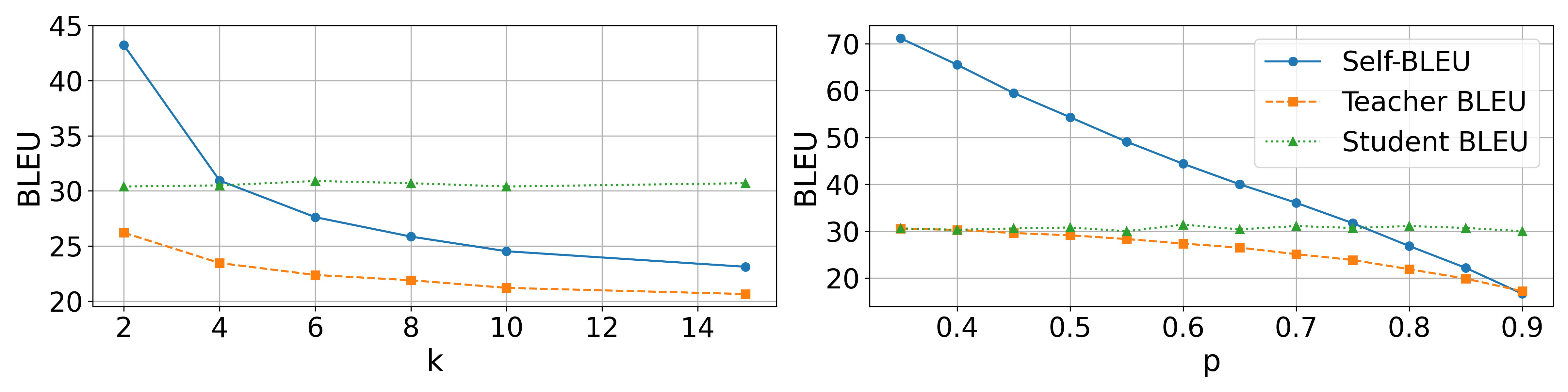}
    \caption{Relationship between the teacher translation quality and variability and the student models score for {\sf eng-swh}. Initial corpus: 100k sentences.}
    \label{fg:params_students}
    \Description{...}
\end{figure}

The adjustment of the sampling parameters affects both output variability and translation quality, making the output more similar to greedy decoding or ancestral sampling.
To gauge the sensitivity of MHD  to the values of $p$ and $k$, we conducted experiments on {\sf eng-swh}, translating 100k {\sf eng} sentences with $M$=10. Fig.~\ref{fg:params_students} illustrates the impact of $p$ and $k$ values on both the translation performance of the student and teacher models (measured by BLEU), and the similarity of the translations (measured by self-BLEU). We use BLEU in this graph for a better comparison with self-BLEU. As observed, higher values of $p$ and $k$ result in more diverse translations, albeit with poorer teacher performance, while maintaining similar performance for the student models.
Finally, we repeated the experiment with 1 million sentences and found that the results were consistent with our previous findings using a smaller corpus. These results suggest that the trade-off between quality and variability is independent of corpus size when using the same decoding method.

\subsection{Gender bias analysis}
\label{bias}

Sequence-level KD typically amplifies biases present in the teacher model due to the over-representation of frequent tokens~\cite{ahn-etal-2022-knowledge}. To assess whether MHD can mitigate this issue, we employed contrastive conditioning~\cite{vamvas-sennrich-2021-contrastive} to evaluate gender bias, using NLLB-200 1.3B as the evaluator model and the WinoMT dataset~\cite{stanovsky-etal-2019-evaluating}.

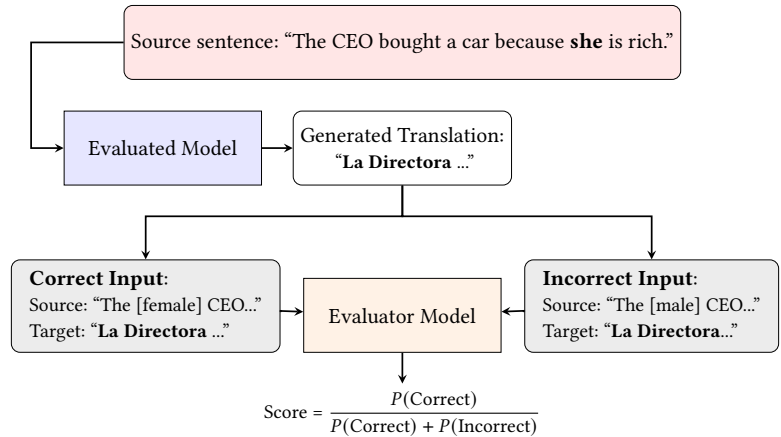
\begin{figure}[tb]
\centering
\scalebox{0.82}{
\begin{tikzpicture}[
    node distance=0.8cm and 1.2cm,
    block/.style={rectangle, draw, minimum width=3.2cm, minimum height=1.2cm, align=center, rounded corners},
    model_block/.style={rectangle, draw, minimum width=3.2cm, minimum height=1.2cm, align=center},
    evalblock/.style={block, fill=gray!15, minimum width=3.5cm},
    arrow/.style={->, thick, >=stealth},
    label/.style={font=\small}
]

\node[block, fill=red!10] (source) at (2,0) {Source sentence: ``The CEO bought a car because \textbf{she} is rich.''};

\node[block, , below=0.5 of source] (translation) {Generated Translation: \\ \small{ ``\textbf{La Directora} ...''}};

\node[model_block, fill=blue!10, left=0.5cm of translation] (model) {Evaluated Model};

\node[evalblock, below left=1.2cm and 0.2cm of translation] (correct-input) {
\begin{tabular}{l}
\textbf{Correct Input}: \\
\small{Source: ``The [female] CEO...''}\\
\small{Target: ``\textbf{La Directora} ...''}
\end{tabular}};

\node[evalblock, below right=1.2cm and 0.2cm of translation] (incorrect-input) {
\begin{tabular}{l}
\textbf{Incorrect Input}: \\
\small{Source: ``The [male] CEO...''}\\
\small{Target: ``\textbf{La Directora}...''}
\end{tabular}};

\node[model_block, fill=orange!10, below=1.5cm of translation] (evaluator) {Evaluator Model};

\node[label, below=0.5cm of evaluator] (score) {
Score = $\displaystyle\frac{P(\text{Correct})}{P(\text{Correct}) + P(\text{Incorrect})}$};

\draw[arrow] (source.west) -- ++(-1.5,0) |- (model.west);
\draw[arrow] (model) -- (translation);
\draw[arrow] (translation.south) -- ++(0,-0.5) -| (correct-input.north);
\draw[arrow] (translation.south) -- ++(0,-0.5) -| (incorrect-input.north);
\draw[arrow] (correct-input) -- node[left] {$.$} (evaluator);
\draw[arrow] (incorrect-input) -- node[right] {$.$} (evaluator);
\draw[arrow] (evaluator) -- (score);

\end{tikzpicture}
}
\caption{Scheme of contrastive conditioning. The evaluator model calculates the probability of the generated translation for both correct and incorrect disambiguated sources. A translation aligned with the correct source will produce a score close to 1, and a translation aligned with the incorrect source will produce a score close to 0.}
\label{fig:bias_evaluation}
\Description{...}
\end{figure}

Contrastive conditioning is a method that leverages the probability assigned by a translation model to a generated translation when presented with controlled variations of the input. Specifically, it involves generating a translation from the original source sentence with the evaluated model and then calculating the probability of the generated translation with the evaluator model when the input is a disambiguated version of the source. If the evaluated model is unbiased, the probability assigned to the translation should be higher when the disambiguated input aligns with the correct gender. Table \ref{tb:cc_example} shows an example of the disambiguated input and the expected output, and Figure \ref{fig:bias_evaluation} illustrates the evaluation protocol.

\begin{table}[tb]
\centering
\resizebox{0.8\columnwidth}{!}{
\begin{tabular}{c|c}

 & \textbf{Example} \\
\hline
Original Source Sentence & The CEO bought a car because \textbf{she} is rich. \\

Correct Disambiguation Cue & The \textbf{[female]} CEO bought a car because \textbf{she} is rich. \\

Incorrect Disambiguation Cue & The \textbf{[male]} CEO bought a car because \textbf{she} is rich. \\

Correct Spanish Translation & \textbf{La Directora General} compró un coche porque es \textbf{rica} \\

Incorrect Spanish Translation & \textbf{El Director General} compró un coche porque es \textbf{rico} \\

\end{tabular}
}
\caption{Contrastive conditioning for gender bias detection. The output of an unbiased model from the original source should match the output of the evaluator model from the correct disambiguation cue.}
\label{tb:cc_example}
\end{table}

Among the languages in this study, WinoMT only contains a dataset for English, so we can only evaluate models translating from English into Swahili, Igbo and Bambara. For each language pair, we evaluated the models trained with \(\mathcal{D}^{1}_{\mathrm{BS}}\) and \(\mathcal{D}^{10}_{Z}\) (\(Z \in \{\mathrm{BS}, \mathrm{DBS}, \text{top-$p$}, \text{top-$k$}, \mathrm{MBR}\}\)). The results in Table \ref{tb:bias} show that generating multiple translations for training reduces gender bias compared to training with a single translation. Although all methods show improvement, sampling-based methods achieve greater bias mitigation by avoiding the over-representation of the most likely tokens inherent in BS.

\begin{table}[tb]
\centering
\resizebox{0.7\columnwidth}{!}{
\begin{tabular}{r|r|r|r|r|r|r|r}
                      & \textbf{NLLB-1.3B} & \(\mathcal{D}^{1}_{\mathrm{BS}}\)  & \(\mathcal{D}^{10}_{\mathrm{BS}}\) & \(\mathcal{D}^{10}_{\mathrm{DBS}}\) & \(\mathcal{D}^{10}_{\text{top-$p$}}\) & \(\mathrm{D}^{10}_{\text{top-$k$}}\) & \(\mathrm{D}^{10}_{\mathrm{MBR}}\)\\ \hline
eng-swh   & 52.9 & 49.2 & 51.0 & 50.4 & \textbf{51.7} & \textbf{51.7}  & 50.1 \\  
eng-ibo   & 52.7 & 49.4 & 50.2 & 50.4 & \textbf{50.5} & \textbf{50.5} & 49.4 \\  
eng-bam   & 58.3 & 50.8 & 50.3 & 51.5 & 52.3 & 51.3 & \textbf{52.5}\\ 
\end{tabular}
}
\caption{Contrastive conditioning accuracy over WinoMT dataset evaluating gender bias. Higher scores are better and the bold scores mark the best student models.}
\label{tb:bias}
\end{table}

\subsection{Hallucinations}
\label{hallucinations}

Hallucination is a well known but atypical issue in MT~\cite{guerreiro-etal-2023-looking}, where the model generates an output that, despite being fluent, is partially or entirely unrelated to the source~\cite{dale-etal-2023-detecting}. Hallucinations can significantly undermine users' confidence in translation models when they occur. Incorporating alternative translations and increasing the variability of the training corpus may improve the student model's fluency in the target language, but not necessarily its translation adequacy. To evaluate this, we analysed the occurrence of hallucinations in the student models.

\begin{figure}[tb]
    \centering
    \includegraphics[scale=0.27]{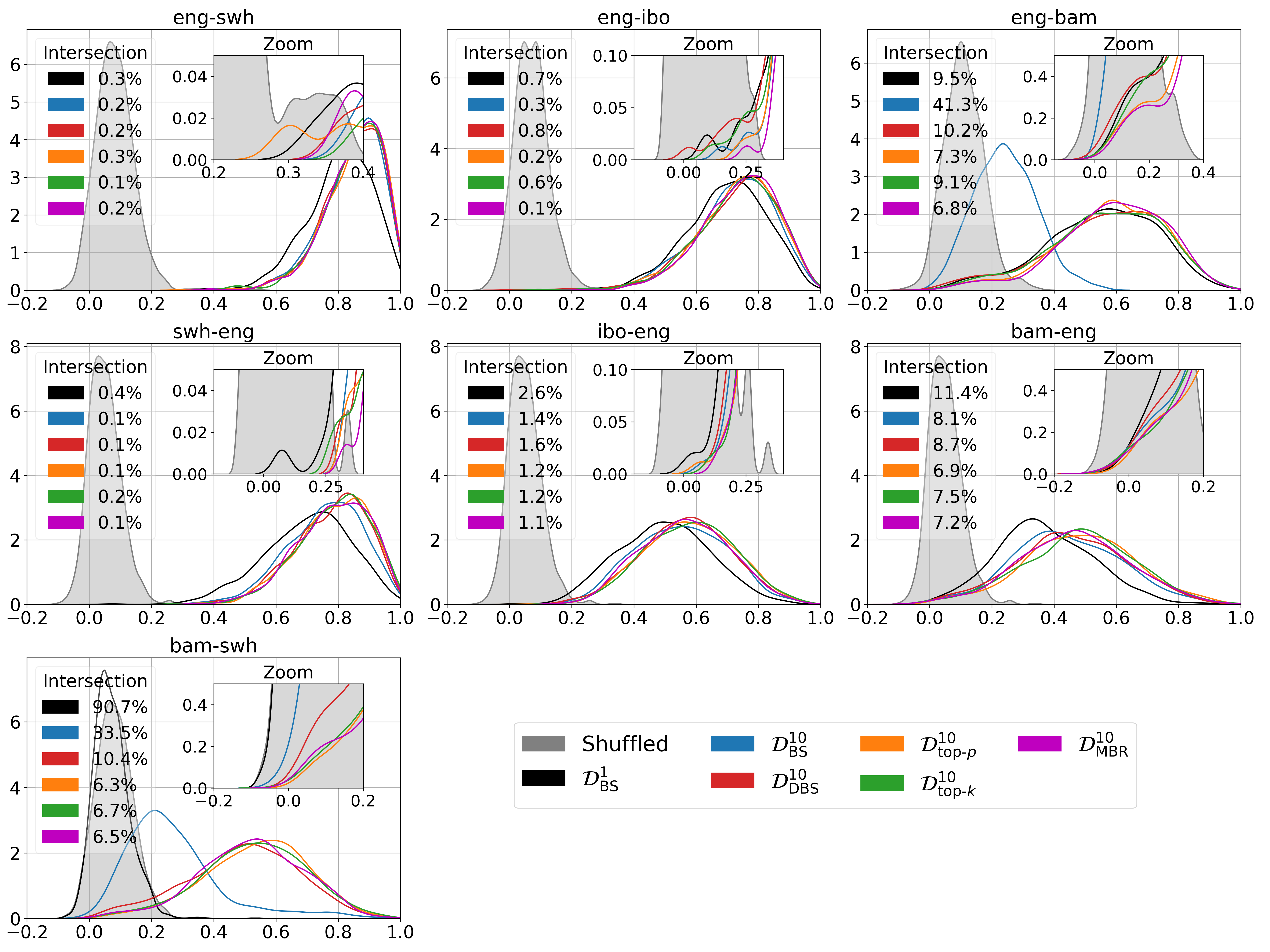}
    \caption{Kernel density estimations (bandwidth=1.0) for SONAR-based cosine similarities between the output produced by student models trained on samples generated with different decoding methods and the reference translations. "Shuffled" denotes the reference sentences shuffled to simulate hallucinations. The intersection highlights the overlap between each model's output and the shuffled area.}
    \label{fg:kde}
    \Description{...}
\end{figure}

Several studies~\cite{dale-etal-2023-detecting, xu-etal-2023-understanding} have shown that using cross-lingual sentence embeddings to measure the similarity between the model output and the reference yields better hallucination detection than metrics such as COMET~\cite{guerreiro-etal-2023-looking}.
Therefore, we computed sentence-level SONAR~\cite{duquenne2023sonar} embeddings for the system outputs and the references, and then computed cosine similarities between them. To find the values of cosine similarity of SONAR embeddings representing hallucinations, we shuffled the references and computed the cosine similarities between the shuffled and original references. Figure \ref{fg:kde} displays the kernel density estimation of the cosine similarity distributions for the systems. The shaded area represents the shuffled references, which cluster near zero, where hallucinations are expected to appear~\cite{dale-etal-2023-detecting}.

In most language pairs, models trained with MHD exhibit fewer hallucinations than those trained with traditional sequence-level KD. The exception is {\sf eng-bam}, where the teacher model performs particularly poor, leading to more hallucinations in student models trained with \(\mathcal{D}^{10}_{\mathrm{BS}}\) compared to \(\mathcal{D}^{1}_{\mathrm{BS}}\). Potentially fewer hallucinations occur in models producing low-resource languages when trained with sampling-based translations than for models trained with deterministic translations, while differences are minimal for models generating English.
This aligns with our observations in Section \ref{sec:100k_sampling} regarding the decline in quality of deterministic methods when generating multiple translations in languages where the teacher is poorly adapted.

\section{Concluding remarks}
\label{discussion}

This study has investigated the effectiveness of MHD, a technique that generates multiple translations from the same source sentence in sequence-level KD, as well as the effect of different decoding methods.
The results show that increasing the number of translations has a positive effect on the student model performance, especially when monolingual data is limited. Using this method, we achieve similar results to standard sequence-level KD with a much smaller monolingual corpus and improve the results with the same corpus size.

MHD matches or slightly outperforms the teacher from English to low-resource languages (scenario from English), but leaves a gap in translation into English. In multilingual models, it may not be possible to extract all the bilingual knowledge from the teacher model with only the synthetic parallel corpus of one language pair, since thanks to transfer learning, part of the translation ability comes from other language pairs. In NLLB-200, which is trained on different parallel corpora with English as the target, a small monolingual Swahili corpus translated into English by the teacher cannot capture all the English knowledge of the model. In this scenario (into English), MHD produces better results than traditional KD with BS (\(\mathcal{D}_{\mathrm{BS}}^{1}\)), but does not improve the performance of the teacher. A similar pattern is observed in the zero-shot scenario, with the key difference being that in this case \(\mathcal{D}^{1}_{\mathrm{BS}}\) fails to train effective student models. In contrast, datasets generated using MHD allow the training of student models that achieve performance close to that of the teacher.

Other effects of increasing variability are that it reduces hallucinations and gender bias. This finding holds for all decoding methods, demonstrating the generalisation capability of the approach. In addition to the overall good results, sampling methods achieve greater mitigation of bias by avoiding the over-representation of the most likely tokens inherent in BS.

Sampling methods allow for a more diverse corpus for learning when generating multiple translations, which is particularly beneficial for low-resource scenarios ({\sf ibo-eng}, {\sf bam-eng}, {\sf bam-swh}). MBR yields the best results for extremely low-resource languages, but it is the slowest method. Top-$p$ is very close in performance, and much faster, so it is preferable in most cases. Nevertheless, with high-resource source languages, the quality of the translations and the mitigation of exposure bias obtained by BS based methods can compensate the low variability of these decoding methods, as occurs with {\sf eng-ibo}, {\sf eng-bam} and {\sf eng-swh}. Especially, when the teacher model contains a lot of knowledge about the source and target languages, it is able to produce multiple translations with a high probability. This explains why DBS gives the best result for {\sf swh-eng} when translating 1 million sentences.

\section*{Ethics Statement}
Knowledge distillation endeavors to produce smaller, more resource-efficient MT systems, thereby diminishing energy requirements compared to the original teacher systems and consequently aiding in the reduction of $\mathrm{CO}_2$ emissions. Moreover, it lowers the entry barrier for deploying MT models, as the resulting models work on lower power hardware. Our student models are remarkably compact, operating at a mere 5\% of the teacher model size.
However, delving into knowledge distillation necessitates a substantial number of training iterations, each accompanied by its own energy consumption. For the experiments detailed in this paper, we trained 482 Transformer models employing NVIDIA GeForce RTX 2080 Ti GPUs. Furthermore, all corpora and tools utilised in this study are available under open source licenses, ensuring the complete reproducibility of the presented results.

\begin{acks}
This paper is part of the R+D+i project PID2021-27999NB-I00 funded by the Spanish Ministry of Science and Innovation (MCIN), the Spanish Research Agency (AEI/10.13039/501100011033) and the European Regional Development Fund A way to make Europe. Some of the computational resources used were funded by the Valencia Government and the European Regional Development Fund (ERDF) through project IDIFEDER/2020/003.
\end{acks}

\bibliographystyle{ACM-Reference-Format}
\bibliography{sample-base,anthology,custom}


\begin{thebibliography}{68}


\ifx \showCODEN    \undefined \def \showCODEN     #1{\unskip}     \fi
\ifx \showDOI      \undefined \def \showDOI       #1{#1}\fi
\ifx \showISBNx    \undefined \def \showISBNx     #1{\unskip}     \fi
\ifx \showISBNxiii \undefined \def \showISBNxiii  #1{\unskip}     \fi
\ifx \showISSN     \undefined \def \showISSN      #1{\unskip}     \fi
\ifx \showLCCN     \undefined \def \showLCCN      #1{\unskip}     \fi
\ifx \shownote     \undefined \def \shownote      #1{#1}          \fi
\ifx \showarticletitle \undefined \def \showarticletitle #1{#1}   \fi
\ifx \showURL      \undefined \def \showURL       {\relax}        \fi
\providecommand\bibfield[2]{#2}
\providecommand\bibinfo[2]{#2}
\providecommand\natexlab[1]{#1}
\providecommand\showeprint[2][]{arXiv:#2}

\bibitem[Adelani et~al\mbox{.}(2022)]%
        {adelani-etal-2022-findings}
\bibfield{author}{\bibinfo{person}{David Adelani}, \bibinfo{person}{Md~Mahfuz~Ibn Alam}, \bibinfo{person}{Antonios Anastasopoulos}, \bibinfo{person}{Akshita Bhagia}, \bibinfo{person}{Marta~R. Costa-juss{\`a}}, \bibinfo{person}{Jesse Dodge}, \bibinfo{person}{Fahim Faisal}, \bibinfo{person}{Christian Federmann}, \bibinfo{person}{Natalia Fedorova}, \bibinfo{person}{Francisco Guzm{\'a}n}, \bibinfo{person}{Sergey Koshelev}, \bibinfo{person}{Jean Maillard}, \bibinfo{person}{Vukosi Marivate}, \bibinfo{person}{Jonathan Mbuya}, \bibinfo{person}{Alexandre Mourachko}, \bibinfo{person}{Safiyyah Saleem}, \bibinfo{person}{Holger Schwenk}, {and} \bibinfo{person}{Guillaume Wenzek}.} \bibinfo{year}{2022}\natexlab{}.
\newblock \showarticletitle{Findings of the {WMT}{'}22 Shared Task on Large-Scale Machine Translation Evaluation for {A}frican Languages}. In \bibinfo{booktitle}{\emph{Proceedings of the Seventh Conference on Machine Translation (WMT)}}, \bibfield{editor}{\bibinfo{person}{Philipp Koehn}, \bibinfo{person}{Lo{\"\i}c Barrault}, \bibinfo{person}{Ond{\v{r}}ej Bojar}, \bibinfo{person}{Fethi Bougares}, \bibinfo{person}{Rajen Chatterjee}, \bibinfo{person}{Marta~R. Costa-juss{\`a}}, \bibinfo{person}{Christian Federmann}, \bibinfo{person}{Mark Fishel}, \bibinfo{person}{Alexander Fraser}, \bibinfo{person}{Markus Freitag}, \bibinfo{person}{Yvette Graham}, \bibinfo{person}{Roman Grundkiewicz}, \bibinfo{person}{Paco Guzman}, \bibinfo{person}{Barry Haddow}, \bibinfo{person}{Matthias Huck}, \bibinfo{person}{Antonio Jimeno~Yepes}, \bibinfo{person}{Tom Kocmi}, \bibinfo{person}{Andr{\'e} Martins}, \bibinfo{person}{Makoto Morishita}, \bibinfo{person}{Christof Monz}, \bibinfo{person}{Masaaki Nagata}, \bibinfo{person}{Toshiaki
  Nakazawa}, \bibinfo{person}{Matteo Negri}, \bibinfo{person}{Aur{\'e}lie N{\'e}v{\'e}ol}, \bibinfo{person}{Mariana Neves}, \bibinfo{person}{Martin Popel}, \bibinfo{person}{Marco Turchi}, {and} \bibinfo{person}{Marcos Zampieri}} (Eds.). \bibinfo{publisher}{Association for Computational Linguistics}, \bibinfo{address}{Abu Dhabi, United Arab Emirates (Hybrid)}, \bibinfo{pages}{773--800}.
\newblock
\urldef\tempurl%
\url{https://aclanthology.org/2022.wmt-1.72}
\showURL{%
\tempurl}


\bibitem[Agarwal et~al\mbox{.}(2024)]%
        {agarwal2024onpolicy}
\bibfield{author}{\bibinfo{person}{Rishabh Agarwal}, \bibinfo{person}{Nino Vieillard}, \bibinfo{person}{Yongchao Zhou}, \bibinfo{person}{Piotr Stanczyk}, \bibinfo{person}{Sabela~Ramos Garea}, \bibinfo{person}{Matthieu Geist}, {and} \bibinfo{person}{Olivier Bachem}.} \bibinfo{year}{2024}\natexlab{}.
\newblock \showarticletitle{On-Policy Distillation of Language Models: Learning from Self-Generated Mistakes}. In \bibinfo{booktitle}{\emph{The Twelfth International Conference on Learning Representations}}.
\newblock
\urldef\tempurl%
\url{https://openreview.net/forum?id=3zKtaqxLhW}
\showURL{%
\tempurl}


\bibitem[Ahn et~al\mbox{.}(2022)]%
        {ahn-etal-2022-knowledge}
\bibfield{author}{\bibinfo{person}{Jaimeen Ahn}, \bibinfo{person}{Hwaran Lee}, \bibinfo{person}{Jinhwa Kim}, {and} \bibinfo{person}{Alice Oh}.} \bibinfo{year}{2022}\natexlab{}.
\newblock \showarticletitle{Why Knowledge Distillation Amplifies Gender Bias and How to Mitigate from the Perspective of {D}istil{BERT}}. In \bibinfo{booktitle}{\emph{Proceedings of the 4th Workshop on Gender Bias in Natural Language Processing (GeBNLP)}}. \bibinfo{publisher}{Association for Computational Linguistics}, \bibinfo{address}{Seattle, Washington}, \bibinfo{pages}{266--272}.
\newblock
\urldef\tempurl%
\url{https://doi.org/10.18653/v1/2022.gebnlp-1.27}
\showDOI{\tempurl}


\bibitem[Basu et~al\mbox{.}(2021)]%
        {basu2021mirostat}
\bibfield{author}{\bibinfo{person}{Sourya Basu}, \bibinfo{person}{Govardana~Sachitanandam Ramachandran}, \bibinfo{person}{Nitish~Shirish Keskar}, {and} \bibinfo{person}{Lav~R. Varshney}.} \bibinfo{year}{2021}\natexlab{}.
\newblock \bibinfo{title}{Mirostat: A Neural Text Decoding Algorithm that Directly Controls Perplexity}.
\newblock
\newblock
\showeprint[arxiv]{2007.14966}~[cs.CL]


\bibitem[Burchell et~al\mbox{.}(2022)]%
        {burchell-birch-and-kenneth-heafield-2022-exploring}
\bibfield{author}{\bibinfo{person}{Laurie Burchell}, \bibinfo{person}{Alexandra Birch}, {and} \bibinfo{person}{Kenneth Heafield}.} \bibinfo{year}{2022}\natexlab{}.
\newblock \showarticletitle{Exploring diversity in back translation for low-resource machine translation}. In \bibinfo{booktitle}{\emph{Proceedings of the Third Workshop on Deep Learning for Low-Resource Natural Language Processing}}. \bibinfo{publisher}{Association for Computational Linguistics}, \bibinfo{address}{Hybrid}, \bibinfo{pages}{67--79}.
\newblock
\urldef\tempurl%
\url{https://doi.org/10.18653/v1/2022.deeplo-1.8}
\showDOI{\tempurl}


\bibitem[Dale et~al\mbox{.}(2023)]%
        {dale-etal-2023-detecting}
\bibfield{author}{\bibinfo{person}{David Dale}, \bibinfo{person}{Elena Voita}, \bibinfo{person}{Loic Barrault}, {and} \bibinfo{person}{Marta~R. Costa-juss{\`a}}.} \bibinfo{year}{2023}\natexlab{}.
\newblock \showarticletitle{Detecting and Mitigating Hallucinations in Machine Translation: Model Internal Workings Alone Do Well, Sentence Similarity {E}ven Better}. In \bibinfo{booktitle}{\emph{Proceedings of the 61st Annual Meeting of the Association for Computational Linguistics (Volume 1: Long Papers)}}, \bibfield{editor}{\bibinfo{person}{Anna Rogers}, \bibinfo{person}{Jordan Boyd-Graber}, {and} \bibinfo{person}{Naoaki Okazaki}} (Eds.). \bibinfo{publisher}{Association for Computational Linguistics}, \bibinfo{address}{Toronto, Canada}, \bibinfo{pages}{36--50}.
\newblock
\urldef\tempurl%
\url{https://doi.org/10.18653/v1/2023.acl-long.3}
\showDOI{\tempurl}


\bibitem[De~Gibert et~al\mbox{.}(2023)]%
        {de-gibert-etal-2023-four}
\bibfield{author}{\bibinfo{person}{Ona De~Gibert}, \bibinfo{person}{Ra{\'u}l V{\'a}zquez}, \bibinfo{person}{Mikko Aulamo}, \bibinfo{person}{Yves Scherrer}, \bibinfo{person}{Sami Virpioja}, {and} \bibinfo{person}{J{\"o}rg Tiedemann}.} \bibinfo{year}{2023}\natexlab{}.
\newblock \showarticletitle{Four Approaches to Low-Resource Multilingual {NMT}: The {H}elsinki Submission to the {A}mericas{NLP} 2023 Shared Task}. In \bibinfo{booktitle}{\emph{Proceedings of the Workshop on Natural Language Processing for Indigenous Languages of the Americas (AmericasNLP)}}, \bibfield{editor}{\bibinfo{person}{Manuel Mager}, \bibinfo{person}{Abteen Ebrahimi}, \bibinfo{person}{Arturo Oncevay}, \bibinfo{person}{Enora Rice}, \bibinfo{person}{Shruti Rijhwani}, \bibinfo{person}{Alexis Palmer}, {and} \bibinfo{person}{Katharina Kann}} (Eds.). \bibinfo{publisher}{Association for Computational Linguistics}, \bibinfo{address}{Toronto, Canada}, \bibinfo{pages}{177--191}.
\newblock
\urldef\tempurl%
\url{https://doi.org/10.18653/v1/2023.americasnlp-1.20}
\showDOI{\tempurl}


\bibitem[DeLucia et~al\mbox{.}(2021)]%
        {delucia-etal-2021-decoding}
\bibfield{author}{\bibinfo{person}{Alexandra DeLucia}, \bibinfo{person}{Aaron Mueller}, \bibinfo{person}{Xiang~Lisa Li}, {and} \bibinfo{person}{Jo{\~a}o Sedoc}.} \bibinfo{year}{2021}\natexlab{}.
\newblock \showarticletitle{Decoding Methods for Neural Narrative Generation}. In \bibinfo{booktitle}{\emph{Proceedings of the 1st Workshop on Natural Language Generation, Evaluation, and Metrics (GEM 2021)}}. \bibinfo{publisher}{Association for Computational Linguistics}, \bibinfo{address}{Online}, \bibinfo{pages}{166--185}.
\newblock
\urldef\tempurl%
\url{https://doi.org/10.18653/v1/2021.gem-1.16}
\showDOI{\tempurl}


\bibitem[Do and Lee(2023)]%
        {10.1145/3546067}
\bibfield{author}{\bibinfo{person}{Heejin Do} {and} \bibinfo{person}{Gary~Geunbae Lee}.} \bibinfo{year}{2023}\natexlab{}.
\newblock \showarticletitle{Target-Oriented Knowledge Distillation with Language-Family-Based Grouping for Multilingual NMT}.
\newblock \bibinfo{journal}{\emph{ACM Trans. Asian Low-Resour. Lang. Inf. Process.}} \bibinfo{volume}{22}, \bibinfo{number}{2}, Article \bibinfo{articleno}{42} (\bibinfo{date}{mar} \bibinfo{year}{2023}), \bibinfo{numpages}{18}~pages.
\newblock
\showISSN{2375-4699}
\urldef\tempurl%
\url{https://doi.org/10.1145/3546067}
\showDOI{\tempurl}


\bibitem[Duquenne et~al\mbox{.}(2023)]%
        {duquenne2023sonar}
\bibfield{author}{\bibinfo{person}{Paul-Ambroise Duquenne}, \bibinfo{person}{Holger Schwenk}, {and} \bibinfo{person}{Benoît Sagot}.} \bibinfo{year}{2023}\natexlab{}.
\newblock \bibinfo{title}{SONAR: Sentence-Level Multimodal and Language-Agnostic Representations}.
\newblock
\newblock
\showeprint[arxiv]{2308.11466}~[cs.CL]
\urldef\tempurl%
\url{https://arxiv.org/abs/2308.11466}
\showURL{%
\tempurl}


\bibitem[Eikema and Aziz(2020)]%
        {eikema-aziz-2020-map}
\bibfield{author}{\bibinfo{person}{Bryan Eikema} {and} \bibinfo{person}{Wilker Aziz}.} \bibinfo{year}{2020}\natexlab{}.
\newblock \showarticletitle{Is {MAP} Decoding All You Need? The Inadequacy of the Mode in Neural Machine Translation}. In \bibinfo{booktitle}{\emph{Proceedings of the 28th International Conference on Computational Linguistics}}. \bibinfo{publisher}{International Committee on Computational Linguistics}, \bibinfo{address}{Barcelona, Spain (Online)}, \bibinfo{pages}{4506--4520}.
\newblock
\urldef\tempurl%
\url{https://doi.org/10.18653/v1/2020.coling-main.398}
\showDOI{\tempurl}


\bibitem[Eikema and Aziz(2022)]%
        {eikema-aziz-2022-sampling}
\bibfield{author}{\bibinfo{person}{Bryan Eikema} {and} \bibinfo{person}{Wilker Aziz}.} \bibinfo{year}{2022}\natexlab{}.
\newblock \showarticletitle{Sampling-Based Approximations to Minimum {B}ayes Risk Decoding for Neural Machine Translation}. In \bibinfo{booktitle}{\emph{Proceedings of the 2022 Conference on Empirical Methods in Natural Language Processing}}, \bibfield{editor}{\bibinfo{person}{Yoav Goldberg}, \bibinfo{person}{Zornitsa Kozareva}, {and} \bibinfo{person}{Yue Zhang}} (Eds.). \bibinfo{publisher}{Association for Computational Linguistics}, \bibinfo{address}{Abu Dhabi, United Arab Emirates}, \bibinfo{pages}{10978--10993}.
\newblock
\urldef\tempurl%
\url{https://doi.org/10.18653/v1/2022.emnlp-main.754}
\showDOI{\tempurl}


\bibitem[Enis and Hopkins(2024)]%
        {enis2024llmnmtadvancinglowresource}
\bibfield{author}{\bibinfo{person}{Maxim Enis} {and} \bibinfo{person}{Mark Hopkins}.} \bibinfo{year}{2024}\natexlab{}.
\newblock \bibinfo{title}{From LLM to NMT: Advancing Low-Resource Machine Translation with Claude}.
\newblock
\newblock
\showeprint[arxiv]{2404.13813}~[cs.CL]
\urldef\tempurl%
\url{https://arxiv.org/abs/2404.13813}
\showURL{%
\tempurl}


\bibitem[Fan et~al\mbox{.}(2021)]%
        {fan-etal-2020-m2m}
\bibfield{author}{\bibinfo{person}{Angela Fan}, \bibinfo{person}{Shruti Bhosale}, \bibinfo{person}{Holger Schwenk}, \bibinfo{person}{Zhiyi Ma}, \bibinfo{person}{Ahmed El-Kishky}, \bibinfo{person}{Siddharth Goyal}, \bibinfo{person}{Mandeep Baines}, \bibinfo{person}{Onur Celebi}, \bibinfo{person}{Guillaume Wenzek}, \bibinfo{person}{Vishrav Chaudhary}, \bibinfo{person}{Naman Goyal}, \bibinfo{person}{Tom Birch}, \bibinfo{person}{Vitaliy Liptchinsky}, \bibinfo{person}{Sergey Edunov}, \bibinfo{person}{Michael Auli}, {and} \bibinfo{person}{Armand Joulin}.} \bibinfo{year}{2021}\natexlab{}.
\newblock \showarticletitle{Beyond {English}-Centric Multilingual Machine Translation}.
\newblock \bibinfo{journal}{\emph{Journal of Machine Learning Research}} \bibinfo{volume}{22}, \bibinfo{number}{107} (\bibinfo{year}{2021}), \bibinfo{pages}{1--48}.
\newblock
\urldef\tempurl%
\url{http://jmlr.org/papers/v22/20-1307.html}
\showURL{%
\tempurl}


\bibitem[Fan et~al\mbox{.}(2018)]%
        {fan2018hierarchical}
\bibfield{author}{\bibinfo{person}{Angela Fan}, \bibinfo{person}{Mike Lewis}, {and} \bibinfo{person}{Yann Dauphin}.} \bibinfo{year}{2018}\natexlab{}.
\newblock \bibinfo{title}{Hierarchical Neural Story Generation}.
\newblock
\newblock
\showeprint[arxiv]{1805.04833}~[cs.CL]


\bibitem[Finkelstein and Freitag(2024)]%
        {finkelstein2024mbr}
\bibfield{author}{\bibinfo{person}{Mara Finkelstein} {and} \bibinfo{person}{Markus Freitag}.} \bibinfo{year}{2024}\natexlab{}.
\newblock \showarticletitle{{MBR} and {QE} Finetuning: Training-time Distillation of the Best and Most Expensive Decoding Methods}. In \bibinfo{booktitle}{\emph{The Twelfth International Conference on Learning Representations}}.
\newblock
\urldef\tempurl%
\url{https://openreview.net/forum?id=bkNx3O0sND}
\showURL{%
\tempurl}


\bibitem[Galiano-Jim{\'e}nez et~al\mbox{.}(2025)]%
        {galiano-jimenez-etal-2025-beyond}
\bibfield{author}{\bibinfo{person}{Aar{\'o}n Galiano-Jim{\'e}nez}, \bibinfo{person}{Juan~Antonio P{\'e}rez-Ortiz}, \bibinfo{person}{Felipe S{\'a}nchez-Mart{\'i}nez}, {and} \bibinfo{person}{V{\'i}ctor~M. S{\'a}nchez-Cartagena}.} \bibinfo{year}{2025}\natexlab{}.
\newblock \showarticletitle{Beyond the Mode: Sequence-Level Distillation of Multilingual Translation Models for Low-Resource Language Pairs}. In \bibinfo{booktitle}{\emph{Findings of the Association for Computational Linguistics: NAACL 2025}}, \bibfield{editor}{\bibinfo{person}{Luis Chiruzzo}, \bibinfo{person}{Alan Ritter}, {and} \bibinfo{person}{Lu~Wang}} (Eds.). \bibinfo{publisher}{Association for Computational Linguistics}, \bibinfo{address}{Albuquerque, New Mexico}, \bibinfo{pages}{6661--6676}.
\newblock
\showISBNx{979-8-89176-195-7}
\urldef\tempurl%
\url{https://doi.org/10.18653/v1/2025.findings-naacl.372}
\showDOI{\tempurl}


\bibitem[Galiano-Jim{\'e}nez et~al\mbox{.}(2023)]%
        {galiano-jimenez-etal-2023-exploiting}
\bibfield{author}{\bibinfo{person}{Aar{\'o}n Galiano-Jim{\'e}nez}, \bibinfo{person}{Felipe S{\'a}nchez-Mart{\'\i}nez}, \bibinfo{person}{V{\'\i}ctor~M. S{\'a}nchez-Cartagena}, {and} \bibinfo{person}{Juan~Antonio P{\'e}rez-Ortiz}.} \bibinfo{year}{2023}\natexlab{}.
\newblock \showarticletitle{Exploiting large pre-trained models for low-resource neural machine translation}. In \bibinfo{booktitle}{\emph{Proceedings of the 24th Annual Conference of the European Association for Machine Translation}}, \bibfield{editor}{\bibinfo{person}{Mary Nurminen}, \bibinfo{person}{Judith Brenner}, \bibinfo{person}{Maarit Koponen}, \bibinfo{person}{Sirkku Latomaa}, \bibinfo{person}{Mikhail Mikhailov}, \bibinfo{person}{Frederike Schierl}, \bibinfo{person}{Tharindu Ranasinghe}, \bibinfo{person}{Eva Vanmassenhove}, \bibinfo{person}{Sergi~Alvarez Vidal}, \bibinfo{person}{Nora Aranberri}, \bibinfo{person}{Mara Nunziatini}, \bibinfo{person}{Carla~Parra Escart{\'\i}n}, \bibinfo{person}{Mikel Forcada}, \bibinfo{person}{Maja Popovic}, \bibinfo{person}{Carolina Scarton}, {and} \bibinfo{person}{Helena Moniz}} (Eds.). \bibinfo{publisher}{European Association for Machine Translation}, \bibinfo{address}{Tampere, Finland}, \bibinfo{pages}{59--68}.
\newblock
\urldef\tempurl%
\url{https://aclanthology.org/2023.eamt-1.7}
\showURL{%
\tempurl}


\bibitem[Goyal et~al\mbox{.}(2020)]%
        {goyal-etal-2020-efficient}
\bibfield{author}{\bibinfo{person}{Vikrant Goyal}, \bibinfo{person}{Sourav Kumar}, {and} \bibinfo{person}{Dipti~Misra Sharma}.} \bibinfo{year}{2020}\natexlab{}.
\newblock \showarticletitle{Efficient Neural Machine Translation for Low-Resource Languages via Exploiting Related Languages}. In \bibinfo{booktitle}{\emph{Proceedings of the 58th Annual Meeting of the Association for Computational Linguistics: Student Research Workshop}}. \bibinfo{publisher}{Association for Computational Linguistics}, \bibinfo{address}{Online}, \bibinfo{pages}{162--168}.
\newblock
\urldef\tempurl%
\url{https://doi.org/10.18653/v1/2020.acl-srw.22}
\showDOI{\tempurl}


\bibitem[Gra{\c{c}}a et~al\mbox{.}(2019)]%
        {graca-etal-2019-generalizing}
\bibfield{author}{\bibinfo{person}{Miguel Gra{\c{c}}a}, \bibinfo{person}{Yunsu Kim}, \bibinfo{person}{Julian Schamper}, \bibinfo{person}{Shahram Khadivi}, {and} \bibinfo{person}{Hermann Ney}.} \bibinfo{year}{2019}\natexlab{}.
\newblock \showarticletitle{Generalizing Back-Translation in Neural Machine Translation}. In \bibinfo{booktitle}{\emph{Proceedings of the Fourth Conference on Machine Translation (Volume 1: Research Papers)}}. \bibinfo{publisher}{Association for Computational Linguistics}, \bibinfo{address}{Florence, Italy}, \bibinfo{pages}{45--52}.
\newblock
\urldef\tempurl%
\url{https://doi.org/10.18653/v1/W19-5205}
\showDOI{\tempurl}


\bibitem[Graves(2012)]%
        {graves2012sequence}
\bibfield{author}{\bibinfo{person}{Alex Graves}.} \bibinfo{year}{2012}\natexlab{}.
\newblock \bibinfo{title}{Sequence Transduction with Recurrent Neural Networks}.
\newblock
\newblock
\showeprint[arxiv]{1211.3711}~[cs.NE]


\bibitem[Guerreiro et~al\mbox{.}(2023)]%
        {guerreiro-etal-2023-looking}
\bibfield{author}{\bibinfo{person}{Nuno~M. Guerreiro}, \bibinfo{person}{Elena Voita}, {and} \bibinfo{person}{Andr{\'e} Martins}.} \bibinfo{year}{2023}\natexlab{}.
\newblock \showarticletitle{Looking for a Needle in a Haystack: A Comprehensive Study of Hallucinations in Neural Machine Translation}. In \bibinfo{booktitle}{\emph{Proceedings of the 17th Conference of the European Chapter of the Association for Computational Linguistics}}, \bibfield{editor}{\bibinfo{person}{Andreas Vlachos} {and} \bibinfo{person}{Isabelle Augenstein}} (Eds.). \bibinfo{publisher}{Association for Computational Linguistics}, \bibinfo{address}{Dubrovnik, Croatia}, \bibinfo{pages}{1059--1075}.
\newblock
\urldef\tempurl%
\url{https://doi.org/10.18653/v1/2023.eacl-main.75}
\showDOI{\tempurl}


\bibitem[Gumma et~al\mbox{.}(2023)]%
        {gumma-etal-2023-empirical}
\bibfield{author}{\bibinfo{person}{Varun Gumma}, \bibinfo{person}{Raj Dabre}, {and} \bibinfo{person}{Pratyush Kumar}.} \bibinfo{year}{2023}\natexlab{}.
\newblock \showarticletitle{An Empirical Study of Leveraging Knowledge Distillation for Compressing Multilingual Neural Machine Translation Models}. In \bibinfo{booktitle}{\emph{Proceedings of the 24th Annual Conference of the European Association for Machine Translation}}, \bibfield{editor}{\bibinfo{person}{Mary Nurminen}, \bibinfo{person}{Judith Brenner}, \bibinfo{person}{Maarit Koponen}, \bibinfo{person}{Sirkku Latomaa}, \bibinfo{person}{Mikhail Mikhailov}, \bibinfo{person}{Frederike Schierl}, \bibinfo{person}{Tharindu Ranasinghe}, \bibinfo{person}{Eva Vanmassenhove}, \bibinfo{person}{Sergi~Alvarez Vidal}, \bibinfo{person}{Nora Aranberri}, \bibinfo{person}{Mara Nunziatini}, \bibinfo{person}{Carla~Parra Escart{\'\i}n}, \bibinfo{person}{Mikel Forcada}, \bibinfo{person}{Maja Popovic}, \bibinfo{person}{Carolina Scarton}, {and} \bibinfo{person}{Helena Moniz}} (Eds.). \bibinfo{publisher}{European Association for Machine Translation}, \bibinfo{address}{Tampere, Finland}, \bibinfo{pages}{103--114}.
\newblock
\urldef\tempurl%
\url{https://aclanthology.org/2023.eamt-1.11}
\showURL{%
\tempurl}


\bibitem[Hewitt et~al\mbox{.}(2022)]%
        {hewitt-etal-2022-truncation}
\bibfield{author}{\bibinfo{person}{John Hewitt}, \bibinfo{person}{Christopher Manning}, {and} \bibinfo{person}{Percy Liang}.} \bibinfo{year}{2022}\natexlab{}.
\newblock \showarticletitle{Truncation Sampling as Language Model Desmoothing}. In \bibinfo{booktitle}{\emph{Findings of the Association for Computational Linguistics: EMNLP 2022}}, \bibfield{editor}{\bibinfo{person}{Yoav Goldberg}, \bibinfo{person}{Zornitsa Kozareva}, {and} \bibinfo{person}{Yue Zhang}} (Eds.). \bibinfo{publisher}{Association for Computational Linguistics}, \bibinfo{address}{Abu Dhabi, United Arab Emirates}, \bibinfo{pages}{3414--3427}.
\newblock
\urldef\tempurl%
\url{https://doi.org/10.18653/v1/2022.findings-emnlp.249}
\showDOI{\tempurl}


\bibitem[Hinton et~al\mbox{.}(2015)]%
        {hinton2015distilling}
\bibfield{author}{\bibinfo{person}{Geoffrey Hinton}, \bibinfo{person}{Oriol Vinyals}, {and} \bibinfo{person}{Jeff Dean}.} \bibinfo{year}{2015}\natexlab{}.
\newblock \bibinfo{title}{Distilling the Knowledge in a Neural Network}.
\newblock
\newblock
\showeprint[arxiv]{1503.02531}~[stat.ML]


\bibitem[Holtzman et~al\mbox{.}(2020)]%
        {holtzman2020curious}
\bibfield{author}{\bibinfo{person}{Ari Holtzman}, \bibinfo{person}{Jan Buys}, \bibinfo{person}{Li Du}, \bibinfo{person}{Maxwell Forbes}, {and} \bibinfo{person}{Yejin Choi}.} \bibinfo{year}{2020}\natexlab{}.
\newblock \bibinfo{title}{The Curious Case of Neural Text Degeneration}.
\newblock
\newblock
\showeprint[arxiv]{1904.09751}~[cs.CL]


\bibitem[Iyer et~al\mbox{.}(2024)]%
        {iyer-etal-2024-quality}
\bibfield{author}{\bibinfo{person}{Vivek Iyer}, \bibinfo{person}{Bhavitvya Malik}, \bibinfo{person}{Pavel Stepachev}, \bibinfo{person}{Pinzhen Chen}, \bibinfo{person}{Barry Haddow}, {and} \bibinfo{person}{Alexandra Birch}.} \bibinfo{year}{2024}\natexlab{}.
\newblock \showarticletitle{Quality or Quantity? On Data Scale and Diversity in Adapting Large Language Models for Low-Resource Translation}. In \bibinfo{booktitle}{\emph{Proceedings of the Ninth Conference on Machine Translation}}, \bibfield{editor}{\bibinfo{person}{Barry Haddow}, \bibinfo{person}{Tom Kocmi}, \bibinfo{person}{Philipp Koehn}, {and} \bibinfo{person}{Christof Monz}} (Eds.). \bibinfo{publisher}{Association for Computational Linguistics}, \bibinfo{address}{Miami, Florida, USA}, \bibinfo{pages}{1393--1409}.
\newblock
\urldef\tempurl%
\url{https://doi.org/10.18653/v1/2024.wmt-1.128}
\showDOI{\tempurl}


\bibitem[Kim and Rush(2016)]%
        {kim-rush-2016-sequence}
\bibfield{author}{\bibinfo{person}{Yoon Kim} {and} \bibinfo{person}{Alexander~M. Rush}.} \bibinfo{year}{2016}\natexlab{}.
\newblock \showarticletitle{Sequence-Level Knowledge Distillation}. In \bibinfo{booktitle}{\emph{Proceedings of the 2016 Conference on Empirical Methods in Natural Language Processing}}. \bibinfo{publisher}{Association for Computational Linguistics}, \bibinfo{address}{Austin, Texas}, \bibinfo{pages}{1317--1327}.
\newblock
\urldef\tempurl%
\url{https://doi.org/10.18653/v1/D16-1139}
\showDOI{\tempurl}


\bibitem[Kingma and Ba(2015)]%
        {adam}
\bibfield{author}{\bibinfo{person}{Diederik~P. Kingma} {and} \bibinfo{person}{Jimmy Ba}.} \bibinfo{year}{2015}\natexlab{}.
\newblock \showarticletitle{Adam: {A} Method for Stochastic Optimization}. In \bibinfo{booktitle}{\emph{3rd International Conference on Learning Representations, {ICLR} 2015, Conference Track Proc.}}
\newblock
\urldef\tempurl%
\url{http://arxiv.org/abs/1412.6980}
\showURL{%
\tempurl}


\bibitem[Kocmi et~al\mbox{.}(2024)]%
        {kocmi-etal-2024-findings}
\bibfield{author}{\bibinfo{person}{Tom Kocmi}, \bibinfo{person}{Eleftherios Avramidis}, \bibinfo{person}{Rachel Bawden}, \bibinfo{person}{Ond{\v{r}}ej Bojar}, \bibinfo{person}{Anton Dvorkovich}, \bibinfo{person}{Christian Federmann}, \bibinfo{person}{Mark Fishel}, \bibinfo{person}{Markus Freitag}, \bibinfo{person}{Thamme Gowda}, \bibinfo{person}{Roman Grundkiewicz}, \bibinfo{person}{Barry Haddow}, \bibinfo{person}{Marzena Karpinska}, \bibinfo{person}{Philipp Koehn}, \bibinfo{person}{Benjamin Marie}, \bibinfo{person}{Christof Monz}, \bibinfo{person}{Kenton Murray}, \bibinfo{person}{Masaaki Nagata}, \bibinfo{person}{Martin Popel}, \bibinfo{person}{Maja Popovi{\'c}}, \bibinfo{person}{Mariya Shmatova}, \bibinfo{person}{Steinth{\'o}r Steingr{\'i}msson}, {and} \bibinfo{person}{Vil{\'e}m Zouhar}.} \bibinfo{year}{2024}\natexlab{}.
\newblock \showarticletitle{Findings of the {WMT}24 General Machine Translation Shared Task: The {LLM} Era Is Here but {MT} Is Not Solved Yet}. In \bibinfo{booktitle}{\emph{Proceedings of the Ninth Conference on Machine Translation}}, \bibfield{editor}{\bibinfo{person}{Barry Haddow}, \bibinfo{person}{Tom Kocmi}, \bibinfo{person}{Philipp Koehn}, {and} \bibinfo{person}{Christof Monz}} (Eds.). \bibinfo{publisher}{Association for Computational Linguistics}, \bibinfo{address}{Miami, Florida, USA}, \bibinfo{pages}{1--46}.
\newblock
\urldef\tempurl%
\url{https://doi.org/10.18653/v1/2024.wmt-1.1}
\showDOI{\tempurl}


\bibitem[Kovacs et~al\mbox{.}(2024)]%
        {kovacs-etal-2024-mitigating}
\bibfield{author}{\bibinfo{person}{Geza Kovacs}, \bibinfo{person}{Daniel Deutsch}, {and} \bibinfo{person}{Markus Freitag}.} \bibinfo{year}{2024}\natexlab{}.
\newblock \showarticletitle{Mitigating Metric Bias in Minimum {B}ayes Risk Decoding}. In \bibinfo{booktitle}{\emph{Proceedings of the Ninth Conference on Machine Translation}}, \bibfield{editor}{\bibinfo{person}{Barry Haddow}, \bibinfo{person}{Tom Kocmi}, \bibinfo{person}{Philipp Koehn}, {and} \bibinfo{person}{Christof Monz}} (Eds.). \bibinfo{publisher}{Association for Computational Linguistics}, \bibinfo{address}{Miami, Florida, USA}, \bibinfo{pages}{1063--1094}.
\newblock
\urldef\tempurl%
\url{https://doi.org/10.18653/v1/2024.wmt-1.109}
\showDOI{\tempurl}


\bibitem[Kudo and Richardson(2018)]%
        {kudo-richardson-2018-sentencepiece}
\bibfield{author}{\bibinfo{person}{Taku Kudo} {and} \bibinfo{person}{John Richardson}.} \bibinfo{year}{2018}\natexlab{}.
\newblock \showarticletitle{{S}entence{P}iece: A simple and language independent subword tokenizer and detokenizer for Neural Text Processing}. In \bibinfo{booktitle}{\emph{Proceedings of the 2018 Conference on Empirical Methods in Natural Language Processing: System Demonstrations}}. \bibinfo{publisher}{Association for Computational Linguistics}, \bibinfo{address}{Brussels, Belgium}, \bibinfo{pages}{66--71}.
\newblock
\urldef\tempurl%
\url{https://doi.org/10.18653/v1/D18-2012}
\showDOI{\tempurl}


\bibitem[Kudugunta et~al\mbox{.}(2023)]%
        {kudugunta2023madlad400}
\bibfield{author}{\bibinfo{person}{Sneha Kudugunta}, \bibinfo{person}{Isaac Caswell}, \bibinfo{person}{Biao Zhang}, \bibinfo{person}{Xavier Garcia}, \bibinfo{person}{Christopher~A. Choquette-Choo}, \bibinfo{person}{Katherine Lee}, \bibinfo{person}{Derrick Xin}, \bibinfo{person}{Aditya Kusupati}, \bibinfo{person}{Romi Stella}, \bibinfo{person}{Ankur Bapna}, {and} \bibinfo{person}{Orhan Firat}.} \bibinfo{year}{2023}\natexlab{}.
\newblock \bibinfo{title}{MADLAD-400: A Multilingual And Document-Level Large Audited Dataset}.
\newblock
\newblock
\showeprint[arxiv]{2309.04662}~[cs.CL]


\bibitem[Kulikov et~al\mbox{.}(2019)]%
        {kulikov-etal-2019-importance}
\bibfield{author}{\bibinfo{person}{Ilia Kulikov}, \bibinfo{person}{Alexander Miller}, \bibinfo{person}{Kyunghyun Cho}, {and} \bibinfo{person}{Jason Weston}.} \bibinfo{year}{2019}\natexlab{}.
\newblock \showarticletitle{Importance of Search and Evaluation Strategies in Neural Dialogue Modeling}. In \bibinfo{booktitle}{\emph{Proceedings of the 12th International Conference on Natural Language Generation}}. \bibinfo{publisher}{Association for Computational Linguistics}, \bibinfo{address}{Tokyo, Japan}, \bibinfo{pages}{76--87}.
\newblock
\urldef\tempurl%
\url{https://doi.org/10.18653/v1/W19-8609}
\showDOI{\tempurl}


\bibitem[Kullback and Leibler(1951)]%
        {Kullback1951}
\bibfield{author}{\bibinfo{person}{Solomon Kullback} {and} \bibinfo{person}{Richard~A Leibler}.} \bibinfo{year}{1951}\natexlab{}.
\newblock \showarticletitle{{On Information and Sufficiency}}.
\newblock \bibinfo{journal}{\emph{The Annals of Mathematical Statistics}} \bibinfo{volume}{22}, \bibinfo{number}{1} (\bibinfo{year}{1951}), \bibinfo{pages}{79--86}.
\newblock


\bibitem[Kumar and Byrne(2004)]%
        {kumar-byrne-2004-minimum}
\bibfield{author}{\bibinfo{person}{Shankar Kumar} {and} \bibinfo{person}{William Byrne}.} \bibinfo{year}{2004}\natexlab{}.
\newblock \showarticletitle{Minimum {B}ayes-Risk Decoding for Statistical Machine Translation}. In \bibinfo{booktitle}{\emph{Proceedings of the Human Language Technology Conference of the North {A}merican Chapter of the Association for Computational Linguistics: {HLT}-{NAACL} 2004}}. \bibinfo{publisher}{Association for Computational Linguistics}, \bibinfo{address}{Boston, Massachusetts, USA}, \bibinfo{pages}{169--176}.
\newblock
\urldef\tempurl%
\url{https://aclanthology.org/N04-1022}
\showURL{%
\tempurl}


\bibitem[Lai et~al\mbox{.}(2021)]%
        {lai-etal-2021-lmu}
\bibfield{author}{\bibinfo{person}{Wen Lai}, \bibinfo{person}{Jind{\v{r}}ich Libovick{\'y}}, {and} \bibinfo{person}{Alexander Fraser}.} \bibinfo{year}{2021}\natexlab{}.
\newblock \showarticletitle{The {LMU} {M}unich System for the {WMT} 2021 Large-Scale Multilingual Machine Translation Shared Task}. In \bibinfo{booktitle}{\emph{Proceedings of the Sixth Conference on Machine Translation}}. \bibinfo{publisher}{Association for Computational Linguistics}, \bibinfo{address}{Online}, \bibinfo{pages}{412--417}.
\newblock
\urldef\tempurl%
\url{https://aclanthology.org/2021.wmt-1.49}
\showURL{%
\tempurl}


\bibitem[Li et~al\mbox{.}(2025)]%
        {li2025ssacometllmsoutperformlearned}
\bibfield{author}{\bibinfo{person}{Senyu Li}, \bibinfo{person}{Jiayi Wang}, \bibinfo{person}{Felermino D. M.~A. Ali}, \bibinfo{person}{Colin Cherry}, \bibinfo{person}{Daniel Deutsch}, \bibinfo{person}{Eleftheria Briakou}, \bibinfo{person}{Rui Sousa-Silva}, \bibinfo{person}{Henrique~Lopes Cardoso}, \bibinfo{person}{Pontus Stenetorp}, {and} \bibinfo{person}{David~Ifeoluwa Adelani}.} \bibinfo{year}{2025}\natexlab{}.
\newblock \bibinfo{title}{SSA-COMET: Do LLMs Outperform Learned Metrics in Evaluating MT for Under-Resourced African Languages?}
\newblock
\newblock
\showeprint[arxiv]{2506.04557}~[cs.CL]
\urldef\tempurl%
\url{https://arxiv.org/abs/2506.04557}
\showURL{%
\tempurl}


\bibitem[M{\"u}ller and Sennrich(2021)]%
        {muller-sennrich-2021-understanding}
\bibfield{author}{\bibinfo{person}{Mathias M{\"u}ller} {and} \bibinfo{person}{Rico Sennrich}.} \bibinfo{year}{2021}\natexlab{}.
\newblock \showarticletitle{Understanding the Properties of Minimum {B}ayes Risk Decoding in Neural Machine Translation}. In \bibinfo{booktitle}{\emph{Proceedings of the 59th Annual Meeting of the Association for Computational Linguistics and the 11th International Joint Conference on Natural Language Processing (Volume 1: Long Papers)}}. \bibinfo{publisher}{Association for Computational Linguistics}, \bibinfo{address}{Online}, \bibinfo{pages}{259--272}.
\newblock
\urldef\tempurl%
\url{https://doi.org/10.18653/v1/2021.acl-long.22}
\showDOI{\tempurl}


\bibitem[{NLLB Team} et~al\mbox{.}(2022)]%
        {nllb}
\bibfield{author}{\bibinfo{person}{{NLLB Team}}, \bibinfo{person}{Marta~R. Costa-jussà}, \bibinfo{person}{James Cross}, \bibinfo{person}{Onur Çelebi}, \bibinfo{person}{Maha Elbayad}, \bibinfo{person}{Kenneth Heafield}, \bibinfo{person}{Kevin Heffernan}, \bibinfo{person}{Elahe Kalbassi}, \bibinfo{person}{Janice Lam}, \bibinfo{person}{Daniel Licht}, \bibinfo{person}{Jean Maillard}, \bibinfo{person}{Anna Sun}, \bibinfo{person}{Skyler Wang}, \bibinfo{person}{Guillaume Wenzek}, \bibinfo{person}{Al Youngblood}, \bibinfo{person}{Bapi Akula}, \bibinfo{person}{Loic Barrault}, \bibinfo{person}{Gabriel~Mejia Gonzalez}, \bibinfo{person}{Prangthip Hansanti}, \bibinfo{person}{John Hoffman}, \bibinfo{person}{Semarley Jarrett}, \bibinfo{person}{Kaushik~Ram Sadagopan}, \bibinfo{person}{Dirk Rowe}, \bibinfo{person}{Shannon Spruit}, \bibinfo{person}{Chau Tran}, \bibinfo{person}{Pierre Andrews}, \bibinfo{person}{Necip~Fazil Ayan}, \bibinfo{person}{Shruti Bhosale}, \bibinfo{person}{Sergey Edunov}, \bibinfo{person}{Angela Fan},
  \bibinfo{person}{Cynthia Gao}, \bibinfo{person}{Vedanuj Goswami}, \bibinfo{person}{Francisco Guzmán}, \bibinfo{person}{Philipp Koehn}, \bibinfo{person}{Alexandre Mourachko}, \bibinfo{person}{Christophe Ropers}, \bibinfo{person}{Safiyyah Saleem}, \bibinfo{person}{Holger Schwenk}, {and} \bibinfo{person}{Jeff Wang}.} \bibinfo{year}{2022}\natexlab{}.
\newblock \bibinfo{title}{No Language Left Behind: Scaling Human-Centered Machine Translation}.
\newblock
\newblock
\urldef\tempurl%
\url{https://doi.org/10.48550/ARXIV.2207.04672}
\showDOI{\tempurl}


\bibitem[Pillutla et~al\mbox{.}(2021)]%
        {pillutla2021mauve}
\bibfield{author}{\bibinfo{person}{Krishna Pillutla}, \bibinfo{person}{Swabha Swayamdipta}, \bibinfo{person}{Rowan Zellers}, \bibinfo{person}{John Thickstun}, \bibinfo{person}{Sean Welleck}, \bibinfo{person}{Yejin Choi}, {and} \bibinfo{person}{Zaid Harchaoui}.} \bibinfo{year}{2021}\natexlab{}.
\newblock \bibinfo{title}{MAUVE: Measuring the Gap Between Neural Text and Human Text using Divergence Frontiers}.
\newblock
\newblock
\showeprint[arxiv]{2102.01454}~[cs.CL]


\bibitem[Popovi{\'c}(2017)]%
        {popovic-2017-chrf}
\bibfield{author}{\bibinfo{person}{Maja Popovi{\'c}}.} \bibinfo{year}{2017}\natexlab{}.
\newblock \showarticletitle{chr{F}++: words helping character n-grams}. In \bibinfo{booktitle}{\emph{Proceedings of the Second Conference on Machine Translation}}. \bibinfo{publisher}{Association for Computational Linguistics}, \bibinfo{address}{Copenhagen, Denmark}, \bibinfo{pages}{612--618}.
\newblock
\urldef\tempurl%
\url{https://doi.org/10.18653/v1/W17-4770}
\showDOI{\tempurl}


\bibitem[Ranzato et~al\mbox{.}(2015)]%
        {Ranzato2015SequenceLT}
\bibfield{author}{\bibinfo{person}{Marc'Aurelio Ranzato}, \bibinfo{person}{Sumit Chopra}, \bibinfo{person}{Michael Auli}, {and} \bibinfo{person}{Wojciech Zaremba}.} \bibinfo{year}{2015}\natexlab{}.
\newblock \showarticletitle{Sequence Level Training with Recurrent Neural Networks}.
\newblock \bibinfo{journal}{\emph{CoRR}}  \bibinfo{volume}{abs/1511.06732} (\bibinfo{year}{2015}).
\newblock
\urldef\tempurl%
\url{https://api.semanticscholar.org/CorpusID:7147309}
\showURL{%
\tempurl}


\bibitem[Rei et~al\mbox{.}(2020)]%
        {rei-etal-2020-comet}
\bibfield{author}{\bibinfo{person}{Ricardo Rei}, \bibinfo{person}{Craig Stewart}, \bibinfo{person}{Ana~C Farinha}, {and} \bibinfo{person}{Alon Lavie}.} \bibinfo{year}{2020}\natexlab{}.
\newblock \showarticletitle{{COMET}: A Neural Framework for {MT} Evaluation}. In \bibinfo{booktitle}{\emph{Proceedings of the 2020 Conference on Empirical Methods in Natural Language Processing (EMNLP)}}. \bibinfo{publisher}{Association for Computational Linguistics}, \bibinfo{address}{Online}, \bibinfo{pages}{2685--2702}.
\newblock
\urldef\tempurl%
\url{https://doi.org/10.18653/v1/2020.emnlp-main.213}
\showDOI{\tempurl}


\bibitem[Riezler and Maxwell(2005)]%
        {riezler-maxwell-2005-pitfalls}
\bibfield{author}{\bibinfo{person}{Stefan Riezler} {and} \bibinfo{person}{John~T. Maxwell}.} \bibinfo{year}{2005}\natexlab{}.
\newblock \showarticletitle{On Some Pitfalls in Automatic Evaluation and Significance Testing for {MT}}. In \bibinfo{booktitle}{\emph{Proceedings of the {ACL} Workshop on Intrinsic and Extrinsic Evaluation Measures for Machine Translation and/or Summarization}}. \bibinfo{publisher}{Association for Computational Linguistics}, \bibinfo{address}{Ann Arbor, Michigan}, \bibinfo{pages}{57--64}.
\newblock
\urldef\tempurl%
\url{https://aclanthology.org/W05-0908}
\showURL{%
\tempurl}


\bibitem[S\'{a}nchez-Cartagena et~al\mbox{.}(2018)]%
        {prompsit:2018:WMT}
\bibfield{author}{\bibinfo{person}{V\'{i}ctor~M. S\'{a}nchez-Cartagena}, \bibinfo{person}{Marta Ba{\~n}\'{o}n}, \bibinfo{person}{Sergio Ortiz-Rojas}, {and} \bibinfo{person}{Gema Ram\'{i}rez-S\'{a}nchez}.} \bibinfo{year}{2018}\natexlab{}.
\newblock \showarticletitle{Prompsit's submission to WMT 2018 Parallel Corpus Filtering shared task}. In \bibinfo{booktitle}{\emph{Proceedings of the Third Conference on Machine Translation, Volume 2: Shared Task Papers}}. \bibinfo{publisher}{Association for Computational Linguistics}, \bibinfo{address}{Brussels, Belgium}.
\newblock


\bibitem[Scalvini et~al\mbox{.}(2025)]%
        {scalvini-etal-2025-rethinking}
\bibfield{author}{\bibinfo{person}{Barbara Scalvini}, \bibinfo{person}{Iben~Nyholm Debess}, \bibinfo{person}{Annika Simonsen}, {and} \bibinfo{person}{Hafsteinn Einarsson}.} \bibinfo{year}{2025}\natexlab{}.
\newblock \showarticletitle{Rethinking Low-Resource {MT:} The Surprising Effectiveness of Fine-Tuned Multilingual Models in the {LLM} Age}. In \bibinfo{booktitle}{\emph{Proceedings of the Joint 25th Nordic Conference on Computational Linguistics and 11th Baltic Conference on Human Language Technologies (NoDaLiDa/Baltic-HLT 2025)}}, \bibfield{editor}{\bibinfo{person}{Richard Johansson} {and} \bibinfo{person}{Sara Stymne}} (Eds.). \bibinfo{publisher}{University of Tartu Library}, \bibinfo{address}{Tallinn, Estonia}, \bibinfo{pages}{609--621}.
\newblock
\showISBNx{978-9908-53-109-0}
\urldef\tempurl%
\url{https://aclanthology.org/2025.nodalida-1.62/}
\showURL{%
\tempurl}


\bibitem[Shi et~al\mbox{.}(2024)]%
        {shi-etal-2024-thorough}
\bibfield{author}{\bibinfo{person}{Chufan Shi}, \bibinfo{person}{Haoran Yang}, \bibinfo{person}{Deng Cai}, \bibinfo{person}{Zhisong Zhang}, \bibinfo{person}{Yifan Wang}, \bibinfo{person}{Yujiu Yang}, {and} \bibinfo{person}{Wai Lam}.} \bibinfo{year}{2024}\natexlab{}.
\newblock \showarticletitle{A Thorough Examination of Decoding Methods in the Era of {LLM}s}. In \bibinfo{booktitle}{\emph{Proceedings of the 2024 Conference on Empirical Methods in Natural Language Processing}}, \bibfield{editor}{\bibinfo{person}{Yaser Al-Onaizan}, \bibinfo{person}{Mohit Bansal}, {and} \bibinfo{person}{Yun-Nung Chen}} (Eds.). \bibinfo{publisher}{Association for Computational Linguistics}, \bibinfo{address}{Miami, Florida, USA}, \bibinfo{pages}{8601--8629}.
\newblock
\urldef\tempurl%
\url{https://doi.org/10.18653/v1/2024.emnlp-main.489}
\showDOI{\tempurl}


\bibitem[Song et~al\mbox{.}(2023)]%
        {song2023letz}
\bibfield{author}{\bibinfo{person}{Yewei Song}, \bibinfo{person}{Saad Ezzini}, \bibinfo{person}{Jacques Klein}, \bibinfo{person}{Tegawende Bissyande}, \bibinfo{person}{Clément Lefebvre}, {and} \bibinfo{person}{Anne Goujon}.} \bibinfo{year}{2023}\natexlab{}.
\newblock \bibinfo{title}{Letz Translate: Low-Resource Machine Translation for Luxembourgish}.
\newblock
\newblock
\showeprint[arxiv]{2303.01347}~[cs.CL]


\bibitem[Stanovsky et~al\mbox{.}(2019)]%
        {stanovsky-etal-2019-evaluating}
\bibfield{author}{\bibinfo{person}{Gabriel Stanovsky}, \bibinfo{person}{Noah~A. Smith}, {and} \bibinfo{person}{Luke Zettlemoyer}.} \bibinfo{year}{2019}\natexlab{}.
\newblock \showarticletitle{Evaluating Gender Bias in Machine Translation}. In \bibinfo{booktitle}{\emph{Proceedings of the 57th Annual Meeting of the Association for Computational Linguistics}}. \bibinfo{publisher}{Association for Computational Linguistics}, \bibinfo{address}{Florence, Italy}, \bibinfo{pages}{1679--1684}.
\newblock
\urldef\tempurl%
\url{https://doi.org/10.18653/v1/P19-1164}
\showDOI{\tempurl}


\bibitem[Su et~al\mbox{.}(2022)]%
        {su2022contrastive}
\bibfield{author}{\bibinfo{person}{Yixuan Su}, \bibinfo{person}{Tian Lan}, \bibinfo{person}{Yan Wang}, \bibinfo{person}{Dani Yogatama}, \bibinfo{person}{Lingpeng Kong}, {and} \bibinfo{person}{Nigel Collier}.} \bibinfo{year}{2022}\natexlab{}.
\newblock \bibinfo{title}{A Contrastive Framework for Neural Text Generation}.
\newblock
\newblock
\showeprint[arxiv]{2202.06417}~[cs.CL]


\bibitem[Tan et~al\mbox{.}(2019)]%
        {tan2018multilingual}
\bibfield{author}{\bibinfo{person}{Xu Tan}, \bibinfo{person}{Yi Ren}, \bibinfo{person}{Di He}, \bibinfo{person}{Tao Qin}, {and} \bibinfo{person}{Tie-Yan Liu}.} \bibinfo{year}{2019}\natexlab{}.
\newblock \showarticletitle{Multilingual Neural Machine Translation with Knowledge Distillation}. In \bibinfo{booktitle}{\emph{Seventh International Conference on Learning Representations}}.
\newblock
\urldef\tempurl%
\url{https://openreview.net/forum?id=S1gUsoR9YX}
\showURL{%
\tempurl}


\bibitem[Tran et~al\mbox{.}(2021)]%
        {facebookwmt21}
\bibfield{author}{\bibinfo{person}{Chau Tran}, \bibinfo{person}{Shruti Bhosale}, \bibinfo{person}{James Cross}, \bibinfo{person}{Philipp Koehn}, \bibinfo{person}{Sergey Edunov}, {and} \bibinfo{person}{Angela Fan}.} \bibinfo{year}{2021}\natexlab{}.
\newblock \showarticletitle{Facebook {AI} {WMT21} News Translation Task Submission}. In \bibinfo{booktitle}{\emph{Proc. of the Sixth Conference on Machine Translation (WMT)}}. \bibinfo{pages}{205--215}.
\newblock


\bibitem[Vamvas and Sennrich(2021)]%
        {vamvas-sennrich-2021-contrastive}
\bibfield{author}{\bibinfo{person}{Jannis Vamvas} {and} \bibinfo{person}{Rico Sennrich}.} \bibinfo{year}{2021}\natexlab{}.
\newblock \showarticletitle{Contrastive Conditioning for Assessing Disambiguation in {MT}: {A} Case Study of Distilled Bias}. In \bibinfo{booktitle}{\emph{Proceedings of the 2021 Conference on Empirical Methods in Natural Language Processing}}. \bibinfo{publisher}{Association for Computational Linguistics}, \bibinfo{address}{Online and Punta Cana, Dominican Republic}, \bibinfo{pages}{10246--10265}.
\newblock
\urldef\tempurl%
\url{https://doi.org/10.18653/v1/2021.emnlp-main.803}
\showDOI{\tempurl}


\bibitem[Vamvas and Sennrich(2024)]%
        {vamvas-sennrich-2024-linear}
\bibfield{author}{\bibinfo{person}{Jannis Vamvas} {and} \bibinfo{person}{Rico Sennrich}.} \bibinfo{year}{2024}\natexlab{}.
\newblock \showarticletitle{Linear-time Minimum {B}ayes Risk Decoding with Reference Aggregation}. In \bibinfo{booktitle}{\emph{Proceedings of the 62nd Annual Meeting of the Association for Computational Linguistics (Volume 2: Short Papers)}}, \bibfield{editor}{\bibinfo{person}{Lun-Wei Ku}, \bibinfo{person}{Andre Martins}, {and} \bibinfo{person}{Vivek Srikumar}} (Eds.). \bibinfo{publisher}{Association for Computational Linguistics}, \bibinfo{address}{Bangkok, Thailand}, \bibinfo{pages}{790--801}.
\newblock
\urldef\tempurl%
\url{https://doi.org/10.18653/v1/2024.acl-short.71}
\showDOI{\tempurl}


\bibitem[Vaswani et~al\mbox{.}(2017)]%
        {vaswani2023attention}
\bibfield{author}{\bibinfo{person}{Ashish Vaswani}, \bibinfo{person}{Noam Shazeer}, \bibinfo{person}{Niki Parmar}, \bibinfo{person}{Jakob Uszkoreit}, \bibinfo{person}{Llion Jones}, \bibinfo{person}{Aidan~N. Gomez}, \bibinfo{person}{\L{}ukasz Kaiser}, {and} \bibinfo{person}{Illia Polosukhin}.} \bibinfo{year}{2017}\natexlab{}.
\newblock \showarticletitle{Attention is all you need}. In \bibinfo{booktitle}{\emph{Proceedings of the 31st International Conference on Neural Information Processing Systems}} (Long Beach, California, USA) \emph{(\bibinfo{series}{NIPS'17})}. \bibinfo{publisher}{Curran Associates Inc.}, \bibinfo{address}{Red Hook, NY, USA}, \bibinfo{pages}{6000–6010}.
\newblock
\showISBNx{9781510860964}


\bibitem[Vijayakumar et~al\mbox{.}(2018)]%
        {diverse_beam_search}
\bibfield{author}{\bibinfo{person}{Ashwin Vijayakumar}, \bibinfo{person}{Michael Cogswell}, \bibinfo{person}{Ramprasaath Selvaraju}, \bibinfo{person}{Qing Sun}, \bibinfo{person}{Stefan Lee}, \bibinfo{person}{David Crandall}, {and} \bibinfo{person}{Dhruv Batra}.} \bibinfo{year}{2018}\natexlab{}.
\newblock \showarticletitle{Diverse Beam Search for Improved Description of Complex Scenes}.
\newblock \bibinfo{journal}{\emph{Proceedings of the AAAI Conference on Artificial Intelligence}} \bibinfo{volume}{32}, \bibinfo{number}{1} (\bibinfo{date}{Apr.} \bibinfo{year}{2018}).
\newblock
\urldef\tempurl%
\url{https://doi.org/10.1609/aaai.v32i1.12340}
\showDOI{\tempurl}


\bibitem[Wang et~al\mbox{.}(2024a)]%
        {wang-etal-2024-afrimte}
\bibfield{author}{\bibinfo{person}{Jiayi Wang}, \bibinfo{person}{David~Ifeoluwa Adelani}, \bibinfo{person}{Sweta Agrawal}, \bibinfo{person}{Marek Masiak}, \bibinfo{person}{Ricardo Rei}, \bibinfo{person}{Eleftheria Briakou}, \bibinfo{person}{Marine Carpuat}, \bibinfo{person}{Xuanli He}, \bibinfo{person}{Sofia Bourhim}, \bibinfo{person}{Andiswa Bukula}, \bibinfo{person}{Muhidin Mohamed}, \bibinfo{person}{Temitayo Olatoye}, \bibinfo{person}{Tosin Adewumi}, \bibinfo{person}{Hamam Mokayed}, \bibinfo{person}{Christine Mwase}, \bibinfo{person}{Wangui Kimotho}, \bibinfo{person}{Foutse Yuehgoh}, \bibinfo{person}{Anuoluwapo Aremu}, \bibinfo{person}{Jessica Ojo}, \bibinfo{person}{Shamsuddeen~Hassan Muhammad}, \bibinfo{person}{Salomey Osei}, \bibinfo{person}{Abdul-Hakeem Omotayo}, \bibinfo{person}{Chiamaka Chukwuneke}, \bibinfo{person}{Perez Ogayo}, \bibinfo{person}{Oumaima Hourrane}, \bibinfo{person}{Salma El~Anigri}, \bibinfo{person}{Lolwethu Ndolela}, \bibinfo{person}{Thabiso Mangwana}, \bibinfo{person}{Shafie~Abdi
  Mohamed}, \bibinfo{person}{Hassan Ayinde}, \bibinfo{person}{Oluwabusayo~Olufunke Awoyomi}, \bibinfo{person}{Lama Alkhaled}, \bibinfo{person}{Sana Al-azzawi}, \bibinfo{person}{Naome~A. Etori}, \bibinfo{person}{Millicent Ochieng}, \bibinfo{person}{Clemencia Siro}, \bibinfo{person}{Njoroge Kiragu}, \bibinfo{person}{Eric Muchiri}, \bibinfo{person}{Wangari Kimotho}, \bibinfo{person}{Lyse~Naomi Wamba~Momo}, \bibinfo{person}{Daud Abolade}, \bibinfo{person}{Simbiat Ajao}, \bibinfo{person}{Iyanuoluwa Shode}, \bibinfo{person}{Ricky Macharm}, \bibinfo{person}{Ruqayya~Nasir Iro}, \bibinfo{person}{Saheed~S. Abdullahi}, \bibinfo{person}{Stephen~E. Moore}, \bibinfo{person}{Bernard Opoku}, \bibinfo{person}{Zainab Akinjobi}, \bibinfo{person}{Abeeb Afolabi}, \bibinfo{person}{Nnaemeka Obiefuna}, \bibinfo{person}{Onyekachi~Raphael Ogbu}, \bibinfo{person}{Sam Ochieng{'}}, \bibinfo{person}{Verrah~Akinyi Otiende}, \bibinfo{person}{Chinedu~Emmanuel Mbonu}, \bibinfo{person}{Sakayo Toadoum~Sari}, \bibinfo{person}{Yao Lu}, {and}
  \bibinfo{person}{Pontus Stenetorp}.} \bibinfo{year}{2024}\natexlab{a}.
\newblock \showarticletitle{{A}fri{MTE} and {A}fri{COMET}: Enhancing {COMET} to Embrace Under-resourced {A}frican Languages}. In \bibinfo{booktitle}{\emph{Proceedings of the 2024 Conference of the North American Chapter of the Association for Computational Linguistics: Human Language Technologies (Volume 1: Long Papers)}}, \bibfield{editor}{\bibinfo{person}{Kevin Duh}, \bibinfo{person}{Helena Gomez}, {and} \bibinfo{person}{Steven Bethard}} (Eds.). \bibinfo{publisher}{Association for Computational Linguistics}, \bibinfo{address}{Mexico City, Mexico}, \bibinfo{pages}{5997--6023}.
\newblock
\urldef\tempurl%
\url{https://doi.org/10.18653/v1/2024.naacl-long.334}
\showDOI{\tempurl}


\bibitem[Wang et~al\mbox{.}(2024b)]%
        {wang2024dontthrowawaydata}
\bibfield{author}{\bibinfo{person}{Jun Wang}, \bibinfo{person}{Eleftheria Briakou}, \bibinfo{person}{Hamid Dadkhahi}, \bibinfo{person}{Rishabh Agarwal}, \bibinfo{person}{Colin Cherry}, {and} \bibinfo{person}{Trevor Cohn}.} \bibinfo{year}{2024}\natexlab{b}.
\newblock \bibinfo{title}{Don't Throw Away Data: Better Sequence Knowledge Distillation}.
\newblock
\newblock
\showeprint[arxiv]{2407.10456}~[cs.CL]
\urldef\tempurl%
\url{https://arxiv.org/abs/2407.10456}
\showURL{%
\tempurl}


\bibitem[Wiher et~al\mbox{.}(2022)]%
        {wiher2022decoding}
\bibfield{author}{\bibinfo{person}{Gian Wiher}, \bibinfo{person}{Clara Meister}, {and} \bibinfo{person}{Ryan Cotterell}.} \bibinfo{year}{2022}\natexlab{}.
\newblock \bibinfo{title}{On Decoding Strategies for Neural Text Generators}.
\newblock
\newblock
\showeprint[arxiv]{2203.15721}~[cs.CL]


\bibitem[Wolf et~al\mbox{.}(2020)]%
        {wolf-etal-2020-transformers}
\bibfield{author}{\bibinfo{person}{Thomas Wolf}, \bibinfo{person}{Lysandre Debut}, \bibinfo{person}{Victor Sanh}, \bibinfo{person}{Julien Chaumond}, \bibinfo{person}{Clement Delangue}, \bibinfo{person}{Anthony Moi}, \bibinfo{person}{Pierric Cistac}, \bibinfo{person}{Tim Rault}, \bibinfo{person}{Remi Louf}, \bibinfo{person}{Morgan Funtowicz}, \bibinfo{person}{Joe Davison}, \bibinfo{person}{Sam Shleifer}, \bibinfo{person}{Patrick von Platen}, \bibinfo{person}{Clara Ma}, \bibinfo{person}{Yacine Jernite}, \bibinfo{person}{Julien Plu}, \bibinfo{person}{Canwen Xu}, \bibinfo{person}{Teven Le~Scao}, \bibinfo{person}{Sylvain Gugger}, \bibinfo{person}{Mariama Drame}, \bibinfo{person}{Quentin Lhoest}, {and} \bibinfo{person}{Alexander Rush}.} \bibinfo{year}{2020}\natexlab{}.
\newblock \showarticletitle{Transformers: State-of-the-Art Natural Language Processing}. In \bibinfo{booktitle}{\emph{Proceedings of the 2020 Conference on Empirical Methods in Natural Language Processing: System Demonstrations}}. \bibinfo{publisher}{Association for Computational Linguistics}, \bibinfo{address}{Online}, \bibinfo{pages}{38--45}.
\newblock
\urldef\tempurl%
\url{https://doi.org/10.18653/v1/2020.emnlp-demos.6}
\showDOI{\tempurl}


\bibitem[Xu et~al\mbox{.}(2023)]%
        {xu-etal-2023-understanding}
\bibfield{author}{\bibinfo{person}{Weijia Xu}, \bibinfo{person}{Sweta Agrawal}, \bibinfo{person}{Eleftheria Briakou}, \bibinfo{person}{Marianna~J. Martindale}, {and} \bibinfo{person}{Marine Carpuat}.} \bibinfo{year}{2023}\natexlab{}.
\newblock \showarticletitle{Understanding and Detecting Hallucinations in Neural Machine Translation via Model Introspection}.
\newblock \bibinfo{journal}{\emph{Transactions of the Association for Computational Linguistics}}  \bibinfo{volume}{11} (\bibinfo{year}{2023}), \bibinfo{pages}{546--564}.
\newblock
\urldef\tempurl%
\url{https://doi.org/10.1162/tacl_a_00563}
\showDOI{\tempurl}


\bibitem[Yu et~al\mbox{.}(2021)]%
        {yu-etal-2021-hw}
\bibfield{author}{\bibinfo{person}{Zhengzhe Yu}, \bibinfo{person}{Daimeng Wei}, \bibinfo{person}{Zongyao Li}, \bibinfo{person}{Hengchao Shang}, \bibinfo{person}{Xiaoyu Chen}, \bibinfo{person}{Zhanglin Wu}, \bibinfo{person}{Jiaxin Guo}, \bibinfo{person}{Minghan Wang}, \bibinfo{person}{Lizhi Lei}, \bibinfo{person}{Min Zhang}, \bibinfo{person}{Hao Yang}, {and} \bibinfo{person}{Ying Qin}.} \bibinfo{year}{2021}\natexlab{}.
\newblock \showarticletitle{{HW}-{TSC}{'}s Participation in the {WMT} 2021 Large-Scale Multilingual Translation Task}. In \bibinfo{booktitle}{\emph{Proceedings of the Sixth Conference on Machine Translation}}. \bibinfo{publisher}{Association for Computational Linguistics}, \bibinfo{address}{Online}, \bibinfo{pages}{456--463}.
\newblock
\urldef\tempurl%
\url{https://aclanthology.org/2021.wmt-1.55}
\showURL{%
\tempurl}


\bibitem[Zhang et~al\mbox{.}(2021)]%
        {zhang-etal-2021-trading}
\bibfield{author}{\bibinfo{person}{Hugh Zhang}, \bibinfo{person}{Daniel Duckworth}, \bibinfo{person}{Daphne Ippolito}, {and} \bibinfo{person}{Arvind Neelakantan}.} \bibinfo{year}{2021}\natexlab{}.
\newblock \showarticletitle{Trading Off Diversity and Quality in Natural Language Generation}. In \bibinfo{booktitle}{\emph{Proceedings of the Workshop on Human Evaluation of NLP Systems (HumEval)}}. \bibinfo{publisher}{Association for Computational Linguistics}, \bibinfo{address}{Online}, \bibinfo{pages}{25--33}.
\newblock
\urldef\tempurl%
\url{https://aclanthology.org/2021.humeval-1.3}
\showURL{%
\tempurl}


\bibitem[Zhang et~al\mbox{.}(2023)]%
        {zhang-etal-2023-towards-understanding}
\bibfield{author}{\bibinfo{person}{Songming Zhang}, \bibinfo{person}{Yunlong Liang}, \bibinfo{person}{Shuaibo Wang}, \bibinfo{person}{Yufeng Chen}, \bibinfo{person}{Wenjuan Han}, \bibinfo{person}{Jian Liu}, {and} \bibinfo{person}{Jinan Xu}.} \bibinfo{year}{2023}\natexlab{}.
\newblock \showarticletitle{Towards Understanding and Improving Knowledge Distillation for Neural Machine Translation}. In \bibinfo{booktitle}{\emph{Proceedings of the 61st Annual Meeting of the Association for Computational Linguistics (Volume 1: Long Papers)}}, \bibfield{editor}{\bibinfo{person}{Anna Rogers}, \bibinfo{person}{Jordan Boyd-Graber}, {and} \bibinfo{person}{Naoaki Okazaki}} (Eds.). \bibinfo{publisher}{Association for Computational Linguistics}, \bibinfo{address}{Toronto, Canada}, \bibinfo{pages}{8062--8079}.
\newblock
\urldef\tempurl%
\url{https://doi.org/10.18653/v1/2023.acl-long.448}
\showDOI{\tempurl}


\bibitem[Zhang et~al\mbox{.}(2020)]%
        {zhang2020niutrans}
\bibfield{author}{\bibinfo{person}{Yuhao Zhang}, \bibinfo{person}{Ziyang Wang}, \bibinfo{person}{Runzhe Cao}, \bibinfo{person}{Binghao Wei}, \bibinfo{person}{Weiqiao Shan}, \bibinfo{person}{Shuhan Zhou}, \bibinfo{person}{Abudurexiti Reheman}, \bibinfo{person}{Tao Zhou}, \bibinfo{person}{Xin Zeng}, \bibinfo{person}{Laohu Wang}, {et~al\mbox{.}}} \bibinfo{year}{2020}\natexlab{}.
\newblock \showarticletitle{The niutrans machine translation systems for wmt20}. In \bibinfo{booktitle}{\emph{Proceedings of the Fifth Conference on Machine Translation}}. \bibinfo{pages}{338--345}.
\newblock


\bibitem[Zhu et~al\mbox{.}(2024)]%
        {zhu-etal-2024-multilingual}
\bibfield{author}{\bibinfo{person}{Wenhao Zhu}, \bibinfo{person}{Hongyi Liu}, \bibinfo{person}{Qingxiu Dong}, \bibinfo{person}{Jingjing Xu}, \bibinfo{person}{Shujian Huang}, \bibinfo{person}{Lingpeng Kong}, \bibinfo{person}{Jiajun Chen}, {and} \bibinfo{person}{Lei Li}.} \bibinfo{year}{2024}\natexlab{}.
\newblock \showarticletitle{Multilingual Machine Translation with Large Language Models: Empirical Results and Analysis}. In \bibinfo{booktitle}{\emph{Findings of the Association for Computational Linguistics: NAACL 2024}}, \bibfield{editor}{\bibinfo{person}{Kevin Duh}, \bibinfo{person}{Helena Gomez}, {and} \bibinfo{person}{Steven Bethard}} (Eds.). \bibinfo{publisher}{Association for Computational Linguistics}, \bibinfo{address}{Mexico City, Mexico}, \bibinfo{pages}{2765--2781}.
\newblock
\urldef\tempurl%
\url{https://doi.org/10.18653/v1/2024.findings-naacl.176}
\showDOI{\tempurl}


\bibitem[Zhu et~al\mbox{.}(2018)]%
        {10.1145/3209978.3210080}
\bibfield{author}{\bibinfo{person}{Yaoming Zhu}, \bibinfo{person}{Sidi Lu}, \bibinfo{person}{Lei Zheng}, \bibinfo{person}{Jiaxian Guo}, \bibinfo{person}{Weinan Zhang}, \bibinfo{person}{Jun Wang}, {and} \bibinfo{person}{Yong Yu}.} \bibinfo{year}{2018}\natexlab{}.
\newblock \showarticletitle{Texygen: A Benchmarking Platform for Text Generation Models}. In \bibinfo{booktitle}{\emph{The 41st International ACM SIGIR Conference on Research \& Development in Information Retrieval}} (Ann Arbor, MI, USA) \emph{(\bibinfo{series}{SIGIR '18})}. \bibinfo{publisher}{Association for Computing Machinery}, \bibinfo{address}{New York, NY, USA}, \bibinfo{pages}{1097–1100}.
\newblock
\showISBNx{9781450356572}
\urldef\tempurl%
\url{https://doi.org/10.1145/3209978.3210080}
\showDOI{\tempurl}


\end{thebibliography}

\appendix

\section{Dataset formalisation}
\label{sec:appendix_form}

This appendix formalises the generation of $M$ hypotheses with each decoding method $Z$ and their integration into the datasets used for MHD. We define each set of $M$ translations per source sentence $x^{i}$ as \(\tilde{\mathcal{Y}}_Z^{i} = \{\tilde{y}^{i,1}, ..., \tilde{y}^{i,M}\}\). Therefore, each dataset containing $N$ parallel sentences is:

\begin{equation}
\mathcal{D}_{\mathrm{Z}}^{M} = \bigcup_{i=1}^{N} \bigcup_{m=1}^{M} \{(x^{i}, \tilde{y}^{i,m})\}
\end{equation}

where \(\tilde{y}^{i,m} \in \tilde{\mathcal{Y}}_{\text{Z}}^{i}\).

\paragraph{Beam Search (BS)}

Beam Search maintains a set of \(n\) partial hypotheses \(\mathcal{H}_t\) at each time step \(t\), expanding them by selecting the most \(n\) probable tokens over the vocabulary $\mathcal{V}$:


\begin{equation}
    \mathcal{H}_{t+1} = \text{max}_n \left( \mathcal{H}_t \times \mathcal{V} \right),
\end{equation}

Where \(\text{max}_n\) selects the \(n\) hypotheses with the highest accumulated probability:

\begin{equation}
    P(y_{1:t} \mid x^{i}) = P(y_{1:t-1} \mid x^{i}) \cdot P(y_t \mid y_{<t}, x^{i}).
\end{equation}

The final set of \(M\) translations is:

\begin{equation}
    \tilde{\mathcal{Y}}_{\text{BS}}^{\,i} = \text{max}_M \left( \mathcal{H}_T \right).
\end{equation}

Note that $n$ must be equal to or greater than $M$.

\paragraph{Diverse Beam Search (DBS)}

Diverse Beam Search divides the beam into \(G\) groups and applies a penalty \(\lambda(y_{1:t})\). Each group $g$ is defined as:


\begin{equation}
    \mathcal{H}_{t+1}^{g} = \text{top-}n_g \left( \mathcal{H}_t^{g} \times \mathcal{V} \right),
\end{equation}

Where the probability of each hypothesis is adjusted with the diversity penalty:

\begin{equation}
    P(y_{1:t} \mid x^{i}) = P(y_{1:t-1} \mid x^{i}) \cdot P(y_t \mid y_{<t}, x^{i}) - \lambda(y_{1:t}).
\end{equation}

The final set of \(M\) translations is:

\begin{equation}
    \tilde{\mathcal{Y}}_{\text{DBS}}^{\,i} = \text{top-}M \left( \bigcup_{g=1}^{G} \mathcal{H}_T^{g} \right).
\end{equation}

\paragraph{Top-$k$ Sampling}

At each time step \(t\), the set of the \(k\) most probable tokens is defined as:

\begin{equation}
    \mathcal{V}_t = \text{max}_k \left( P(y_t \mid y_{<t}, x^{i}) \right).
\end{equation}

A translation is generated by sampling from the renormalised distribution over \(\mathcal{V}_t\), called \(P_{\mathcal{V}_t}\):
\begin{equation}
    y_t \sim P_{\mathcal{V}_t}(y_t \mid y_{<t}, x^{i}),
\end{equation}


The set of \(M\) generated translations is:

\begin{equation}
    \tilde{\mathcal{Y}}_{\text{top-}k}^{\,i} = \left\{ \tilde{y}^{i,m} \mid \tilde{y}^{i,m} = \{ y_t \sim \mathcal{V}_t \}_{t=1}^{T} \right\}_{m=1}^{M}.
\end{equation}

\paragraph{Top-$p$}

The set of tokens whose cumulative probability mass reaches \(p\) is defined as:

\begin{equation}
    \mathcal{V}_t = \left\{ y_t \in \text{argmin}_{\mathcal{V}} \sum_{y \in \mathcal{V}} P(y \mid y_{<t}, x^{i}) \geq p \right\}.
\end{equation}

Where argmin returns the smallest set of tokens with a probability mass of $p$. A translation is generated by sampling from the renormalised distribution over \(\mathcal{V}_t\):
\begin{equation}
    y_t \sim P_{\mathcal{V}_t}(y_t \mid y_{<t}, x^{i}),
\end{equation}


The set of translations is:

\begin{equation}
    \tilde{\mathcal{Y}}_{\text{top-}p}^{\,i} = \left\{ \tilde{y}^{i,m} \mid \tilde{y}^{i,m} = \{ y_t \sim \mathcal{V}_t \}_{t=1}^{T} \right\}_{m=1}^{M}.
\end{equation}

\paragraph{Minimum Bayes Risk (MBR)}

We first generate a set of \( n \) hypotheses \(\mathcal{H}(x^{i}) = \{h_1, h_2, \dots, h_n\}\) using epsilon sampling~\cite{hewitt-etal-2022-truncation}. The utility function \( U(h, c) \) measures the similarity between a hypotheses \( h \) and a candidate \( c \). In this case, we computed the utility using fastChrF~\cite{vamvas-sennrich-2024-linear}. The expected utility for a hypothesis \( h \) is defined as:

\begin{equation}
U(h) = \sum_{c \in \mathcal{H}(x^{(i)})} P(c \mid x^{i}) \cdot U(h, c),
\end{equation}

where \( P(c \mid x^{i}) \) is the probability assigned to candidate \( c \) by the teacher model. The optimal translation is the one that maximises the expected utility:

\begin{equation}
\tilde{y}^{i} = \text{argmax}_{h \in \mathcal{H}(x^{i})} U(h).
\end{equation}

To generate \( M \) diverse hypotheses, we select the top \( M \) hypotheses with the highest expected utility:

\begin{equation}
\tilde{\mathcal{Y}}_{\text{MBR}}^{\,i} = \text{max}_M \left( U(h) \mid h \in \mathcal{H}(x^{i}) \right).
\end{equation}

\section{Student models}
\label{sec:appendix_A_2}

Each student model consist of a transformer \cite{vaswani2023attention} with 6 layers for both the encoder and the decoder, embedding dimension of 512, feed-forward inner-layer dimension of 2048, and 8 attention heads. All our models were trained using the \texttt{Fairseq} toolkit\footnote{\url{https://github.com/facebookresearch/fairseq}} and a different joint bilingual SentencePiece~\cite{kudo-richardson-2018-sentencepiece} model for each language pair, trained on the training samples generated from the teacher with a vocabulary of 10,000 tokens. For training we used a learning rate of 0.0007 with the Adam~\cite{adam} optimizer ($\beta_1$=$0.9$, $\beta_2$=$0.98$), 8,000 warm-up updates and 8,000 max tokens. We used dropout of 0.1 and updated the model after 2 training steps. The cross-entropy loss with label smoothing was computed on the development set after every epoch and the best checkpoint was selected after 6 validation steps with no improvement.

\section{Corpora} \label{sec:appendix_A_1}
The largest corpora correspond to English and Swahili. The English corpus is a fragment of OSCAR-3301 dataset\footnote{\url{https://huggingface.co/datasets/oscar-corpus/OSCAR-2301}} and for Swahili we used Monolingual African Languages from ParaCrawls, a collection of corpora available for the joint task Large-Scale Machine Translation Evaluation for African Languages"at WMT22~\cite{adelani-etal-2022-findings}. The Igbo corpus was obtained from the same collection.

To clean these three corpora, we used \texttt{monocleaner}~\cite{prompsit:2018:WMT}. We used the available ready-to-use language packages for English and Swahili and trained a model for Igbo using the Igbo part of the wmt22\_african dataset.\footnote{\url{https://huggingface.co/datasets/allenai/wmt22_african}}
We removed all sentences with a \texttt{monocleaner} score lower than 0.5 and, for English and Swahili, we then randomly picked one million sentences. For Igbo, our final corpus comprises 451,789 sentences.

For Bambara we collected all available corpora in Hugging Face.\footnote{\url {https://huggingface.co/datasets/RobotsMaliAI/bayelemabaga}, \url {https://github.com/masakhane-io/lafand-mt}, \url {https://wortschatz.uni-leipzig.de/en/download/Bambara}, \url {https://github.com/facebookresearch/flores/tree/main/nllb_seed}, \url {https://huggingface.co/datasets/bigscience/xP3}, \url {https://huggingface.co/datasets/allenai/MADLAD-400}} For the MADLAD-400~\cite{kudugunta2023madlad400} corpus we used only the clean part. After concatenating these corpora, we removed duplicated sentences and the result was 108,187 sentences.

\section{Additional results}
\label{sec:appendix_B}
This sections reports additional results to measure the effect of decoding methods in sequence-level KD.

\subsection{NLLB-200 1.3 translation quality}
\label{sec:nllb_1_BLEU}

Table \ref{tb:nllb_bleu_sampling} shows the impact of each decoding method on the translation quality of NLLB-200 1.3B, evaluated with BLEU.
\begin{table}[tb]
\centering
\resizebox{\columnwidth}{!}{
\begin{tabular}{lccccccc} 
\hline
 \small{\textbf{Method}} & \small{\textbf{eng-swh}} & \small{\textbf{eng-ibo}} & \small{\textbf{eng-bam}} & \small{\textbf{swh-eng}} & \small{\textbf{ibo-eng}} & \small{\textbf{bam-eng}} & \small{\textbf{bam-swh}} \\  
\hline
\small{BS}      & \small{$33.1$} & \small{$16.1$} & \small{$\phantom{0}6.8$} & \small{$42.9$} & \small{$30.3$} & \small{$17.8$} & \small{$11.6$} \\  
\small{DBS}     & \small{$32.2$} & \small{$13.7$} & \small{$\phantom{0}6.1$} & \small{$42.4$} & \small{$30.3$} & \small{$16.2$} & \small{$\phantom{0}8.3$} \\  
\small{top-$p$} & \small{$25.1$} & \small{$12.4$} & \small{$\phantom{0}4.9$} & \small{$34.8$} & \small{$24.3$} & \small{$14.4$} & \small{$\phantom{0}8.9$} \\  
\small{top-$k$} & \small{$21.5$} & \small{$10.9$} & \small{$\phantom{0}4.6$} & \small{$28.8$} & \small{$20.9$} & \small{$12.6$} & \small{$\phantom{0}8.3$} \\  
\small{MBR}     & \small{$30.3$} & \small{$15.1$} & \small{$\phantom{0}6.3$} & \small{$42.2$} & \small{$29.8$} & \small{$14.9$} & \small{$10.1$} \\  
\hline
\end{tabular}
}
\caption{BLEU scores of NLLB-200 1.3B on the FLORES+ devtest dataset for different decoding methods: beam search (BS), diverse beam search (DBS), top-$p$ (average of 3 runs), top-$k$ (average of 3 runs), and MBR. The NLLB-200 3.3B results, which show the same relative order between decoding methods, can be found in Appendix \ref{nllb_3_methods}.}
\label{tb:nllb_bleu_sampling}
\end{table}

\subsection{NLLB-200 3.3 translation quality}
\label{nllb_3_methods}

Tables \ref{tb:nllb_3_chrF_sampling} and \ref{tb:nllb_3_bleu_sampling} show the impact of each decoding method on the translation quality of NLLB-200 3.3B, as evaluated using chrF++ and BLEU, respectively.

\begin{table}[tb]
\centering
\resizebox{\columnwidth}{!}{
\begin{tabular}{l|ccccccc} 
 & \textbf{eng-swh} & \textbf{eng-ibo} & \textbf{eng-bam} & \textbf{swh-eng} & \textbf{ibo-eng} & \textbf{bam-eng} & \textbf{bam-swh} \\  
\hline
BS      &  59.2 & 41.2 & 31.1 & 65.0 & 53.3 & 39.0 & 36.0 \\  
DBS     &  59.0 & 40.6 & 29.4 & 64.6 & 52.7 & 38.9 & 35.2 \\  
top-$p$ &  55.9 & 38.6 & 28.5 & 61.1 & 49.7 & 36.8 & 33.7 \\  
top-$k$ &  50.4 & 35.5 & 27.0 & 54.8 & 45.3 & 34.6 & 32.0 \\  
\hline
\end{tabular}
}
\caption{chrF++ scores of NLLB-200 3.3B on the FLORES+ devtest dataset when decoding with beam search, diverse beam search, top-$p$ (average of 3 runs) and top-$k$ (average of 3 runs).}
\label{tb:nllb_3_chrF_sampling}

\end{table}

\begin{table}[tb]
\centering
\resizebox{\columnwidth}{!}{
\begin{tabular}{l|ccccccc} 
 & \textbf{eng-swh} & \textbf{eng-ibo} & \textbf{eng-bam} & \textbf{swh-eng} & \textbf{ibo-eng} & \textbf{bam-eng} & \textbf{bam-swh} \\  
\hline
BS      & $33.9$ & $16.2$ & $\phantom{0}7.0$ & $44.8$ & $32.0$ & $17.5$ & $10.8$ \\  
DBS     & $32.7$ & $15.9$ & $\phantom{0}6.0$ & $44.2$ & $31.1$ & $17.1$ & $10.7$ \\  
top-$p$ & $28.9$ & $14.0$ & $\phantom{0}5.9$ & $39.7$ & $28.1$ & $15.1$ & $\phantom{0}9.3$ \\  
top-$k$ & $22.1$ & $10.8$ & $\phantom{0}4.7$ & $30.9$ & $21.9$ & $12.1$ & $\phantom{0}7.0$ \\  
\hline
\end{tabular}
}
\caption{BLEU scores of NLLB-200 3.3B on the FLORES+ devtest dataset when decoding with beam search, diverse beam search, top-$p$ (average of 3 runs) and top-$k$ (average of 3 runs).}
\label{tb:nllb_3_bleu_sampling}

\end{table}





\subsection{Experiments with 100k sentences}
\label{sec:appendix_B_1}

\begin{figure}[tb]
    \centering
    \includegraphics[scale=0.34]{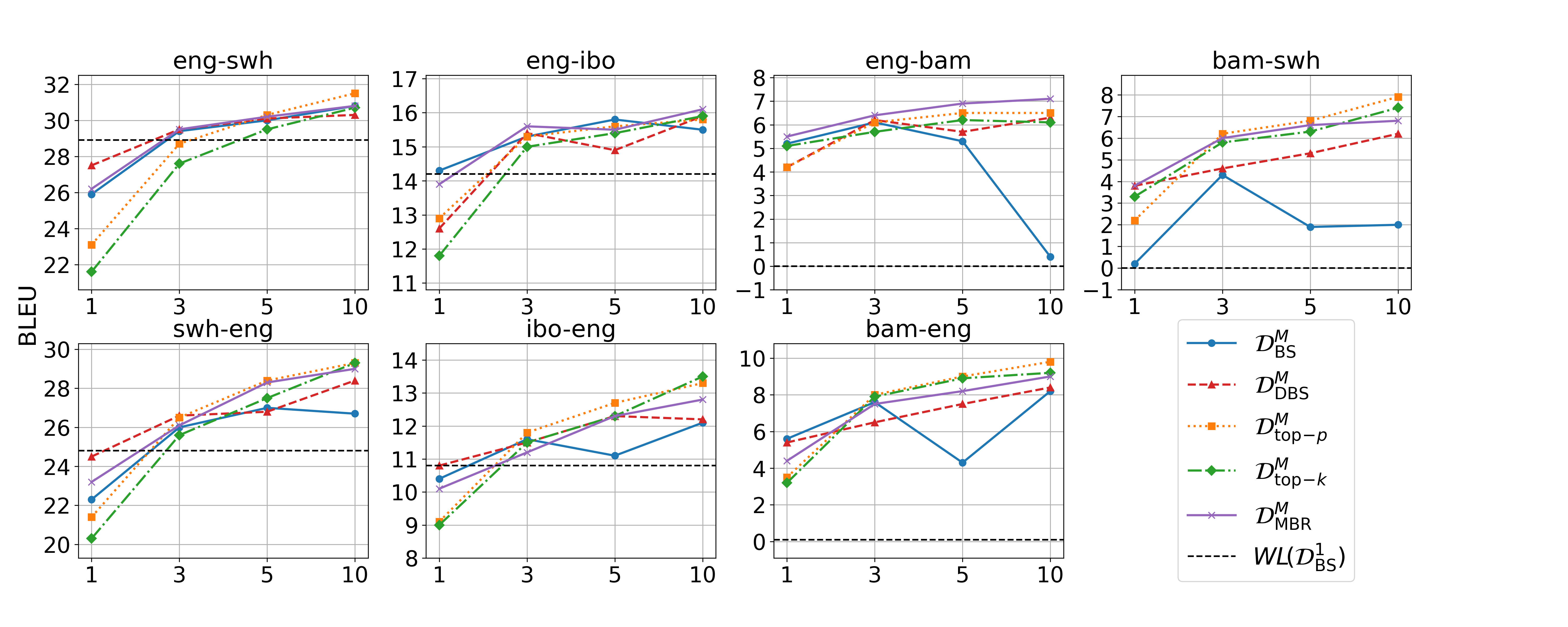}
    \caption{Average BLEU score obtained by student models trained on M samples generated with different decoding methods (x-axis).}  
    \label{fg:bleu_student}
    \Description{MHD results with 100k monolingual sentences and different M values.}
\end{figure}

Figure \ref{fg:bleu_student} shows the BLEU scores obtained by student models trained on corpus generated by NLLB-200 1.3B .

\begin{figure}[tb]
    \centering
    \includegraphics[scale=0.30]{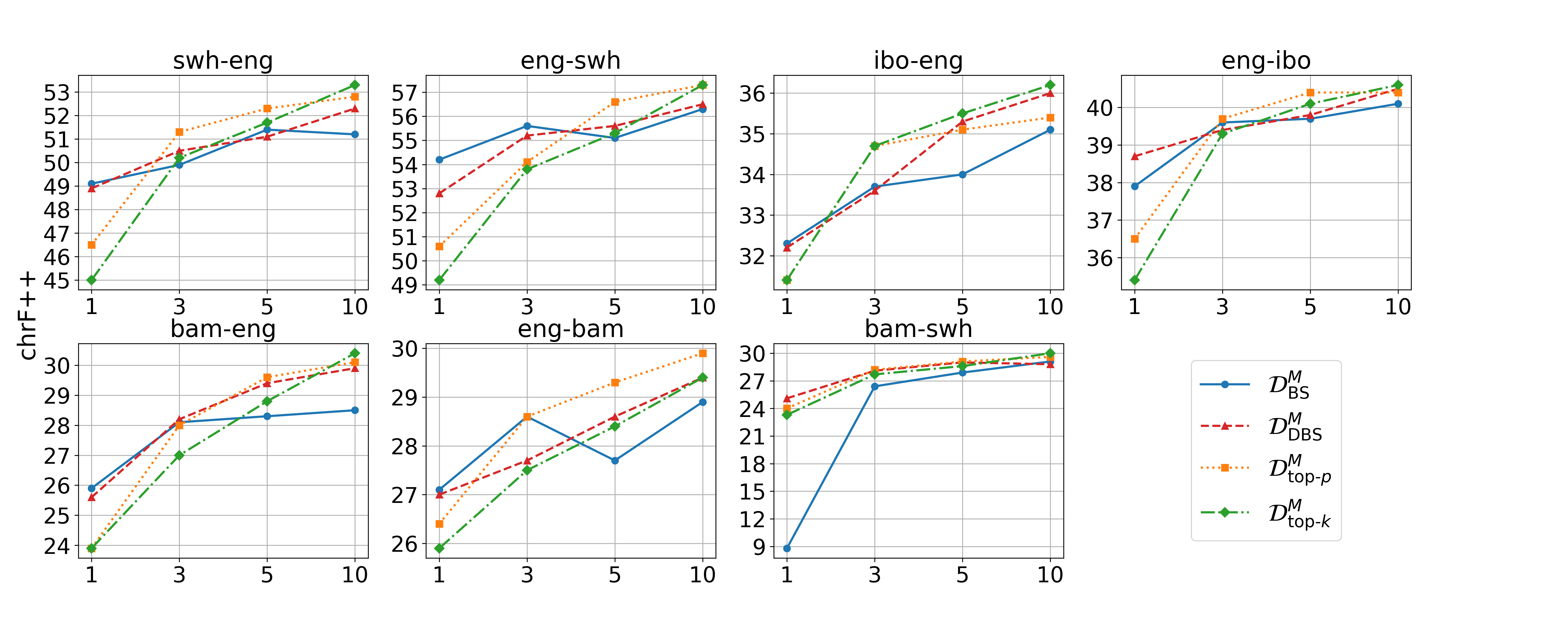}
    \caption{Average chrF++ score obtained by student models trained on M samples generated by NLLB-200 3.3B with different decoding methods (x-axis).}
    \label{fg:students_3.3_chrf}
\end{figure}

\begin{figure}[tb]
    \centering
    \includegraphics[scale=0.30]{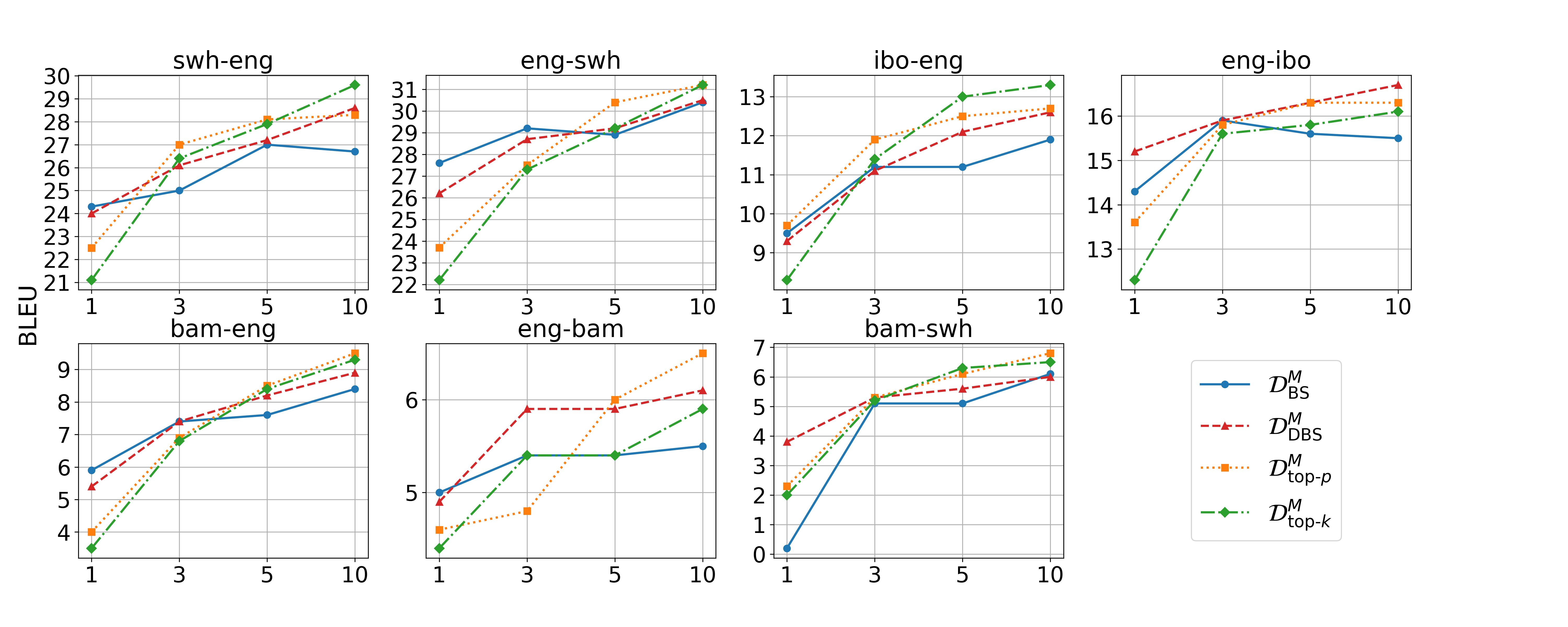}
    \caption{Average BLEU score obtained by student models trained on M samples generated by NLLB-200 3.3B with different decoding methods (x-axis).}
    \label{fg:students_3.3_bleu}
\end{figure}

The results obtained using NLLB-200 3.3B are shown in Figures \ref{fg:students_3.3_chrf} and \ref{fg:students_3.3_bleu}. 

\subsection{Experiments with 500k and 1 million sentences}
\label{sec:appendix_B_2}
Tables \ref{tb:bleu_1M} and \ref{tb:chrf_1M} show the BLEU and chrF++ scores of the trained student models together with the teacher score. The results in Table \ref{tb:chrf_1M} correspond to those in Figure \ref{fg:bleu_student_100k-1M}, together with the directions for which the available corpus does not reach one million sentences.

\begin{table*}[tb]
\centering
\resizebox{\columnwidth}{!}{
\begin{tabular}{l|rrrrrrr}
\textbf{}                      & \small{eng-swh} & \small{eng-ibo} & \small{eng-bam} & \small{swh-eng} & \small{ibo-eng} & \small{bam-eng} & \small{bam-swh} \\ \hline
\textbf{\small{NLLB 1.3B}}                                  &   33.1  &   16.1  &   6.8   &   42.9  &   30.7  &   17.8  &  11.6   \\   \hline

\textbf{\small{Word-level \(\mathcal{D}_{\mathrm{BS}}^{1}\)}}     &   28.9  &   14.2  &   0.0   &   24.8  &   10.8  &   0.1  &  0.0 \\

\textbf{100k \(\mathcal{D}_{\mathrm{BS}}^{1}\)}                     &  26.2   &  14.1   &   4.7   &  22.9   &   10.0  &  5.8    &   2.1  \\   
\textbf{100k \(\mathcal{D}_{\mathrm{BS}}^{10}\)}        &   30.4 &   \underline{\textbf{15.6}} &   0.3  &  27.9  &   12.1 &   8.7  &   1.2 \\   
\textbf{100k \(\mathcal{D}_{\mathrm{DBS}}^{10}\)}        &  \textbf{30.2}  &  \underline{\textbf{15.9}}  &  6.3   &  28.4  &  \textbf{12.4}  &  \textbf{8.3}   &  6.1  \\   
\textbf{100k} \(\mathcal{D}_{\text{top-$p$}}^{10}\)   &   31.1  &   \underline{\textbf{16.0}}  &   6.6   &  29.4   &   13.4  &   9.9   &  7.7 \\   
\textbf{100k} \(\mathcal{D}_{\text{top-$k$}}^{10}\)   &  \textbf{30.9}   &   \underline{\textbf{15.8}}  &   6.3   &  29.1   &   13.4  &   9.7   &   7.5  \\
\textbf{100k}  \(\mathcal{D}_{\mathrm{MBR}}^{10}\)   &  30.8   &   16.1  &   7.1   &  28.9   &   12.8  &   9.0   &   6.8  \\   \hline

\textbf{500k \(\mathcal{D}_{\mathrm{BS}}^{1}\)}         &   31.5  &   15.4  &   6.3   &  31.3   &   14.2  &  --     &   --  \\   
\textbf{500k \(\mathcal{D}_{\mathrm{BS}}^{10}\)}       &   33.5  &   15.1  &   6.4   &  32.7   &   15.5  &  --     &   --  \\   
\textbf{500k \(\mathcal{D}_{\mathrm{DBS}}^{10}\)}      &   33.2  &   16.9  &   6.6   &  34.0   &   16.9  &   --    &   --  \\   
\textbf{500k} \(\mathcal{D}_{\text{top-$p$}}^{10}\)  &   33.9  &    16.9 &   6.9   &  33.3   &   16.2  &  --     &   --  \\   
\textbf{500k} \(\mathcal{D}_{\text{top-$k$}}^{10}\)  &   33.4  &   16.0  &   6.7   &  33.6   &   17.1  &  --     &   --  \\   \hline

\textbf{1M \(\mathcal{D}_{\mathrm{BS}}^{1}\)}                   &   33.5  &   16.5  &   6.5   &  34.0   &   --    &   --    &   --  \\   
\textbf{1M \(\mathcal{D}_{\mathrm{BS}}^{10}\)}      &   34.1  &   17.0  &   6.8   &  34.5   &   --    &   --    &   --  \\   
\textbf{1M \(\mathcal{D}_{\mathrm{DBS}}^{10}\)}      &   34.0  &   15.7  &    6.6  &  35.6  &   --    &   --    &   --  \\   
\textbf{1M} \(\mathcal{D}_{\text{top-$p$}}^{10}\) &    33.8 &   16.6  &   7.1   &  34.1   &  --     &   --    &   --  \\   
\textbf{1M} \(\mathcal{D}_{\text{top-$k$}}^{10}\) &   33.7  &    16.7 &   7.1   &  34.4   &   --    &   --    &   --  \\   \hline

\end{tabular}
}
\caption{BLEU scores on the FLORES+ devtest for several student models and the teacher. Underlined results are those that show no statistically significant difference compared to \(\mathcal{D}_{\mathrm{BS}}^{1}\). Bolded results are those that show no statistically significant difference compared to \(\mathcal{D}_{\mathrm{BS}}^{10}\).}
\label{tb:bleu_1M}
\end{table*}

\begin{table*}[tb]
\centering
\resizebox{\columnwidth}{!}{
\begin{tabular}{l|rrrrrrr}
\textbf{}                      & \small{eng-swh} & \small{eng-ibo} & \small{eng-bam} & \small{swh-eng} & \small{ibo-eng} & \small{bam-eng} & \small{bam-swh} \\ \hline
\textbf{\small{NLLB 1.3B}}     & 59.2 & 41.0 & 30.9 & 63.5 & 52.6 & 38.6 & 35.6 \\  \hline

\textbf{\small{Word-level \(\mathcal{D}_{\mathrm{BS}}^{1}\)}}     &  55.4   &   38.4  &   4.8   &  48.9   &  33.5   &  6.3   &  5.1   \\

\textbf{100k \(\mathcal{D}_{\mathrm{BS}}^{1}\)}         &  52.4   &   37.6  &   27.8   &  47.5   &  32.7   &  25.5   &   8.7  \\

\textbf{100k \(\mathcal{D}_{\mathrm{BS}}^{10}\)}        &  56.9   &   39.6  &   12.6   &  50.7   &  35.5   &  29.1   &  20.4  \\   
\textbf{100k \(\mathcal{D}_{\mathrm{DBS}}^{10}\)}       &  56.6   &   40.0  &   28.9   &  52.3   &  36.0   &  29.5   &  28.9  \\   
\textbf{100k} \(\mathcal{D}_{\text{top-$p$}}^{10}\)     &  57.4   &   39.9  &   30.1   &  52.7   &  36.7   &  30.6   &  30.9  \\   
\textbf{100k} \(\mathcal{D}_{\text{top-$k$}}^{10}\)     &  56.9   &   39.9  &   29.4   &  52.7   &  36.7   &  30.6   &  31.0  \\   
\textbf{100k} \(\mathcal{D}_{\mathrm{MBR}}^{10}\)       &  57.7   &   41.2  &   31.3   &  53.3   &  36.6   &  31.5   &  32.3  \\  \hline

\textbf{\small{500k $\mathcal{D}_{\mathrm{BS}}^{1}$}}       & 57.8 & 40.0 & 30.4 & 54.6 & 37.3 & - & - \\
\textbf{\small{500k $\mathcal{D}_{\mathrm{BS}}^{10}$}}      & 59.2 & 40.1 & 30.7 & 55.7 & 39.2 & - & - \\
\textbf{\small{500k $\mathcal{D}_{\mathrm{DBS}}^{10}$}}     & 59.2 & 41.1 & 29.8 & 56.5 & 40.7 & - & - \\
\textbf{\small{500k $\mathcal{D}_{\text{top}\!-\!p}^{10}$}} & 59.4 & 41.2 & 30.9 & 56.2 & 39.7 & - & - \\
\textbf{\small{500k $\mathcal{D}_{\text{top}\!-\!k}^{10}$}} & 59.2 & 40.5 & 30.9 & 56.7 & 39.8 & - & - \\
\hline
\textbf{\small{1M $\mathcal{D}_{\mathrm{BS}}^{1}$}} & 57.8 & 41.5 & 30.5 & 55.8 & - & - & - \\
\textbf{\small{1M $\mathcal{D}_{\mathrm{BS}}^{10}$}}& 59.7 & 41.6 & 31.0 & 57.0 & - & - & - \\
\textbf{\small{1M $\mathcal{D}_{\mathrm{DBS}}^{10}$}}  & 59.6 & 40.4 & 30.2 & 57.9 & - & - & - \\
\textbf{\small{1M $\mathcal{D}_{\text{top}\!-\!p}^{10}$}} & 59.3 & 41.0 & 31.0 & 56.8 & - & - & - \\
\textbf{\small{1M $\mathcal{D}_{\text{top}\!-\!k}^{10}$}} & 59.4 & 41.1 & 31.0 & 57.1 & - & - & - \\
\hline



\end{tabular}
}
\caption{chrF++ scores on the FLORES+ devtest for several student models and the teacher model.}
\label{tb:chrf_1M}
\end{table*}


\end{document}